\documentclass[preprint,12pt]{elsarticle}

\usepackage{amssymb}
\usepackage{amsmath}

\usepackage{graphicx}
\usepackage{tabularx} % add in preamble

\usepackage[hidelinks]{hyperref}

\usepackage{subcaption} % add this in the preamble

\journal{}

\begin{document}

\begin{frontmatter}



\title{An End-to-End Deep Learning Framework for Arsenicosis Diagnosis Using Mobile-Captured Skin Images}


\author[a]{Asif Newaz\corref{cor1}} %% Author name

\affiliation[a]{organization={Islamic University of Technology (IUT)},%Department and Organization
            addressline={Board Bazar}, 
            city={Gazipur},
            %postcode={}, 
            %state={},
            country={Bangladesh}}

\author[b]{Asif Ur Rahman Adib} %% Author name

\affiliation[b]{organization={Ahsanullah University of Science and Technology},%Department and Organization
            addressline={Tejgaon}, 
            city={Dhaka},
            %postcode={}, 
            %state={},
            country={Bangladesh}}            

\author[a]{Rajit Sahil} %% Author name
\author[a]{Mashfique Mehzad} %% Author name

%% Corresponding author info
\cortext[cor1]{Corresponding author: Asif Newaz, 
Email: \href{mailto:eee.asifnewaz@iut-dhaka.edu}{eee.asifnewaz@iut-dhaka.edu}}

%% Abstract
\begin{abstract}

\textbf{Background:} Arsenicosis is a serious public health concern in South and Southeast Asia, primarily caused by long-term consumption of arsenic-contaminated water. Its early cutaneous manifestations are clinically significant but often underdiagnosed, particularly in rural areas with limited access to dermatologists. Automated, image-based diagnostic solutions can support early detection and timely interventions.

\textbf{Methods:} In this study, we propose an end-to-end framework for arsenicosis diagnosis using mobile phone–captured skin images. A dataset comprising 20 classes and over 11000 images of arsenic-induced and other dermatological conditions was curated. Multiple deep learning architectures, including convolutional neural networks (CNNs) and Transformer-based models, were benchmarked for arsenicosis detection. Model interpretability was integrated via LIME and Grad-CAM, while deployment feasibility was demonstrated through a web-based diagnostic tool.

\textbf{Results:} Transformer-based models significantly outperformed CNNs, with the Swin Transformer achieving the best results (86\% accuracy). LIME and Grad-CAM visualizations confirmed that the models attended to lesion-relevant regions, increasing clinical transparency and aiding in error analysis. The framework also demonstrated strong performance on external validation samples, confirming its ability to generalize beyond the curated dataset.

\textbf{Conclusion:} The proposed framework demonstrates the potential of deep learning for non-invasive, accessible, and explainable diagnosis of arsenicosis from mobile-acquired images. By enabling reliable image-based screening, it can serve as a practical diagnostic aid in rural and resource-limited communities, where access to dermatologists is scarce, thereby supporting early detection and timely intervention.

\end{abstract}

%end{graphicalabstract}

\begin{highlights}

%\item An end-to-end framework for arsenicosis diagnosis is proposed.

\item A dataset containing over 11000 mobile-captured skin lesion images was curated.

\item CNN and Transformer models were benchmarked for arsenicosis detection.

\item Swin Transformer achieved the best performance with 86\% accuracy.

\item LIME and Grad-CAM visual explanations were utilized for model interpretability.

\item External validation with unseen images demonstrated strong generalization across conditions.

\item A web-based diagnostic tool was developed for arsenic screening in rural communities.

\end{highlights}

%% Keywords

\end{frontmatter}


%%%% main text
%%

\section{Introduction}

Skin diseases are among the most common health problems worldwide, impacting millions of individuals across diverse populations. Globally, skin diseases represent a massive public health burden: recent estimates indicate that in 2021 alone, approximately 4.69 billion new cases were recorded, and skin conditions remain the fourth leading cause of non-fatal disease burden and among the top causes of disability worldwide \cite{ILDSSkin2024}. These diseases range from mild infections to chronic and life-threatening disorders. Beyond their high prevalence, skin diseases impose a substantial healthcare and socioeconomic burden, often leading to reduced quality of life, stigmatization, and, in severe cases, long-term complications if left untreated \cite{seth2017global}. Accurate diagnosis remains a major challenge due to the visual similarity among different conditions, which can result in misclassification even by trained experts. These challenges are further exacerbated in resource-limited and rural settings, where access to specialized dermatological care and advanced diagnostic tools is scarce, underscoring the urgent need for reliable, accessible, and cost-effective diagnostic solutions \cite{urban2021burden}.

\begin{figure}[htbp]
    \centering
    \includegraphics[width=\linewidth]{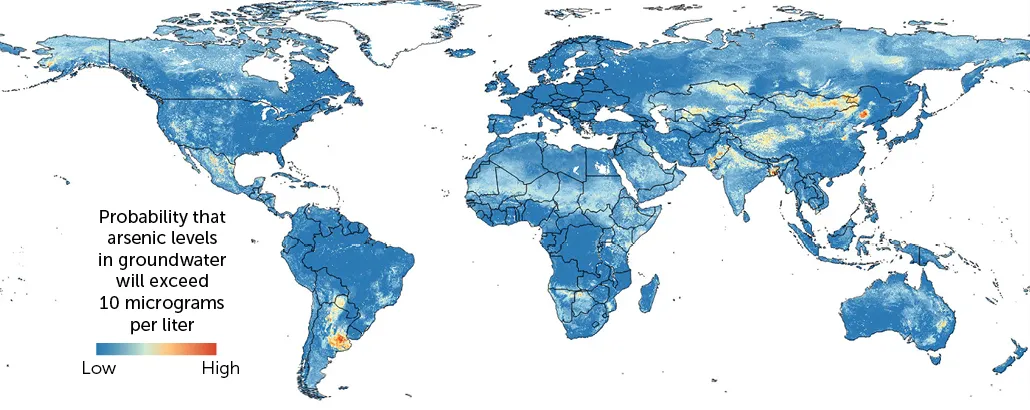}
    \caption{Global prediction of arsenic contamination in groundwater. Areas shown in red indicate a high probability of arsenic levels exceeding the WHO guideline of 10~µg/L, 
    while blue indicates lower risk. Adapted from Podgorski et al. ~\cite{podgorski2020global}}
    \label{fig:arsenic_map}
\end{figure}

One particularly alarming skin-related public health crisis is the widespread outbreak of arsenic-induced skin lesions in South and Southeast Asia. It is also reported in parts of Latin America, North America, and Europe. The global map in Figure \ref{fig:arsenic_map} shows the level of arsenic contamination worldwide. It has been estimated that around 94 to 220 million people worldwide may be exposed to high arsenic levels in groundwater \cite{podgorski2020global}. Since the early 1990s, arsenic contamination in groundwater has emerged as a major environmental and health threat. In countries like Bangladesh, India, and Nepal, millions of tube wells—initially deemed safe—were later found to contain arsenic concentrations far exceeding the WHO's safety threshold of 10 parts per billion (ppb) \cite{WHO2017}. 
In Bangladesh alone, it is estimated that 20–40 million people have been chronically exposed to arsenic-contaminated water \cite{ahmad2018arsenic}. 

%including cardiovascular diseases, diabetes, and cancers of the skin, lung, and bladder. Dermatologically, it leads to cutaneous manifestations such as melanosis, leucomelanosis, keratosis, Bowen’s disease, basal cell carcinoma, and squamous cell carcinoma \cite{Khatun2023}. 

Chronic arsenic exposure is associated with a wide spectrum of systemic and dermatological health consequences. Systemically, long-term ingestion of arsenic-contaminated water has been linked to cardiovascular disorders such as hypertension and peripheral vascular disease, metabolic diseases including diabetes, neurological impairments, and an increased risk of internal cancers, particularly of the lung, bladder, kidney, and liver~\cite{WHOArsenic}. Dermatologically, arsenic toxicity produces some of the most characteristic manifestations of chronic exposure. These include diffuse or spotted hyperpigmentation (melanosis), hypopigmented areas interspersed with dark patches (leucomelanosis), and the development of keratotic lesions on the palms and soles. Over time, these premalignant lesions can progress to Bowen’s disease and invasive skin cancers such as basal cell carcinoma (BCC) and squamous cell carcinoma (SCC)~\cite{Khatun2023}. 

%The presence of these distinctive cutaneous features often serves as an early clinical marker of chronic arsenic poisoning, especially in highly exposed populations.

\begin{figure}[htbp]
    \centering
    \begin{subfigure}[b]{0.48\textwidth}
        \centering
        \includegraphics[width=\linewidth]{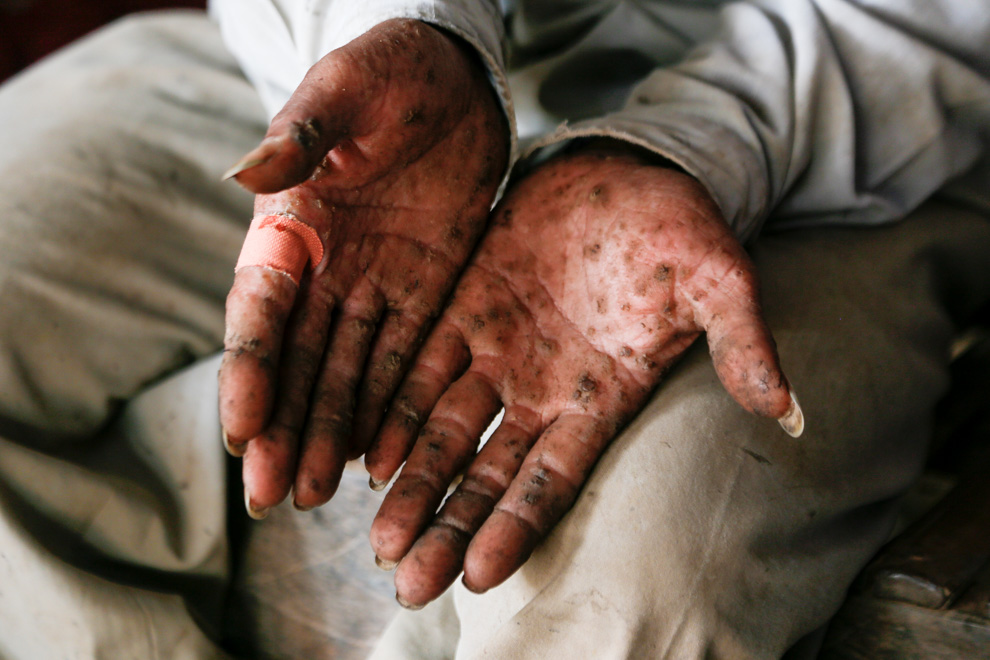}
        %\caption{Palmar keratosis}
        %\label{fig:arsenic_palmar}
    \end{subfigure}
    \hfill
    \begin{subfigure}[b]{0.48\textwidth}
        \centering
        \includegraphics[width=\linewidth]{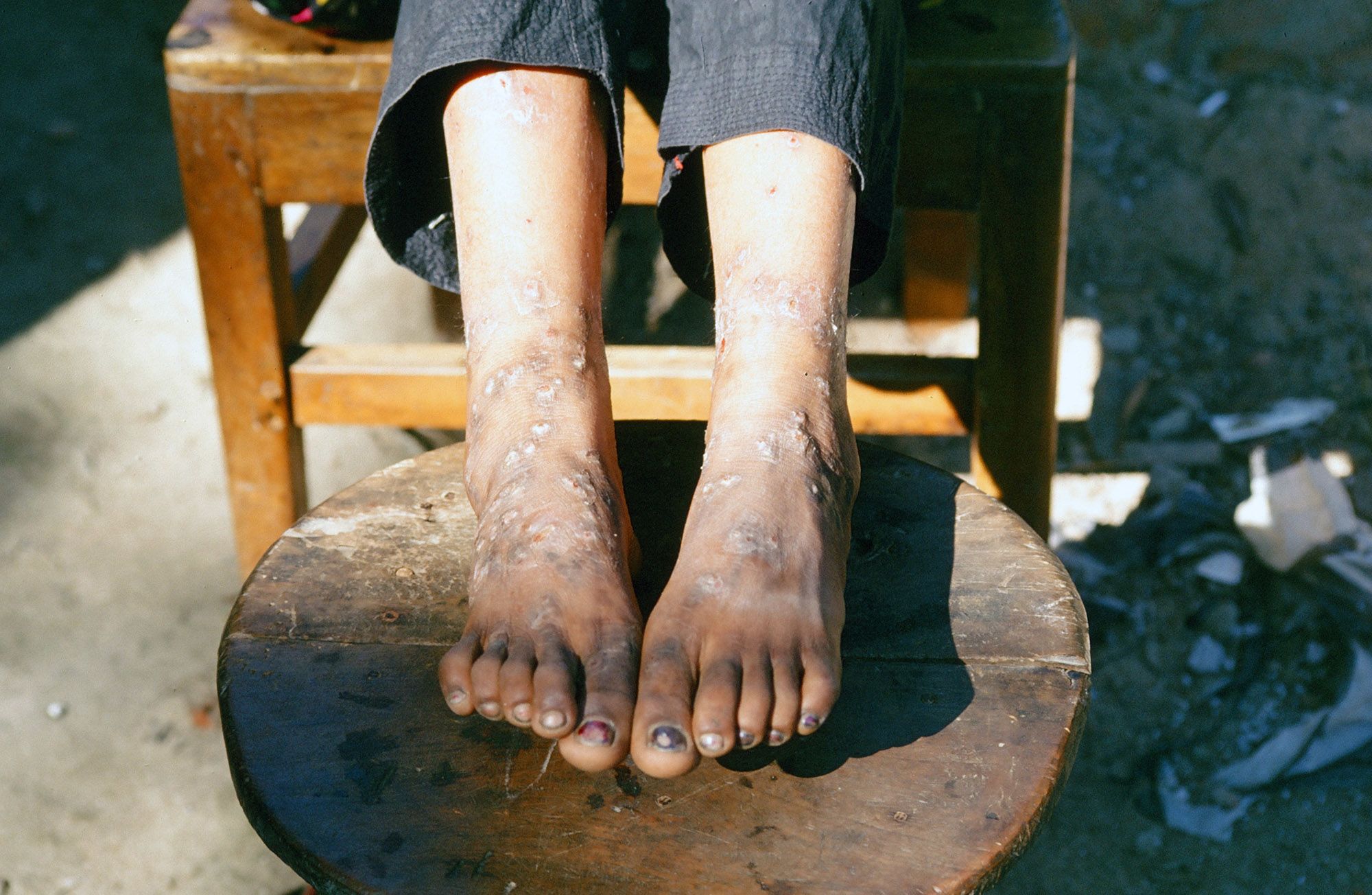}
        %\caption{Plantar keratosis}
        %\label{fig:arsenic_plantar}
    \end{subfigure}
    \caption{Representative examples of arsenic-induced skin lesions.}
    \label{fig:arsenic_samples}
\end{figure}

Arsenical keratosis is one of the earliest and most characteristic cutaneous manifestations of chronic arsenic exposure and is considered a sensitive clinical biomarker of arsenic toxicity~\cite{StatPearls2024}. These skin conditions often appear in multiple areas of the body, may resemble other common dermatoses, and can progress from premalignant lesions to invasive skin cancers, making early and accurate diagnosis crucial to prevent severe complications. Figure \ref{fig:arsenic_samples}. represents two representative examples of arsenic-induced skin manifestations.

Traditional diagnostic workflows for arsenicosis rely on measuring arsenic concentrations in biological matrices such as urine, hair, nails, and blood \cite{WHOArsenic, Abernathy2001}. While these methods are clinically validated, they are invasive, costly, and logistically demanding. Urinary arsenic is generally regarded as the most reliable biomarker for recent exposure and internal dose; however, it is subject to variation due to dietary intake (e.g., seafood arsenicals), age, gender, and other environmental contamination \cite{Vahter2008}. In contrast, hair and nail reflect longer-term exposure but are susceptible to external contamination, which may artificially elevate levels. Blood arsenic concentrations fluctuate substantially due to rapid clearance from circulation, thereby limiting their diagnostic value~\cite{NRC1999}. Moreover, these methods require laboratory-grade instrumentation and trained personnel, which are rarely available in resource-constrained and rural regions. This underscores the urgent need for alternative, accessible, and non-invasive diagnostic strategies.

In this context, visual diagnosis using skin lesion imagery offers a practical alternative for arsenic screening. Dermatologists typically rely on clinical visual inspection as the first step in identifying cutaneous manifestations, occasionally supported by laboratory-based techniques such as biopsies or microscopic examination. However, manual diagnosis is not only labor-intensive and prone to error due to variations in skin tone and lesion appearance, but also largely inaccessible in rural communities. Specialized dermatological expertise is extremely limited in rural regions, where the disease burden is highest. 
These challenges have stimulated increasing interest in automated diagnostic approaches that exploit advances in artificial intelligence (AI), particularly deep learning (DL), to facilitate dermatological screening \cite{esteva2019guide, newaz2022intelligent}. Recent progress in computer-aided diagnostic (CAD) systems has demonstrated the potential of medical image analysis as a non-invasive, cost-effective, and scalable solution for the early detection of skin lesions \cite{chan2020deep}.

With recent advances in DL, it has become feasible to develop robust diagnostic systems that automatically extract and analyze visual features from skin lesion images. Deep neural networks (DNNs), particularly convolutional neural networks (CNNs), have consistently demonstrated high accuracy in the classification of various skin diseases, such as melanoma, eczema, and psoriasis ~\cite{hasan2023survey}. More recently, transformer-based architectures, initially introduced for natural language processing, have been successfully adapted to computer vision tasks. By leveraging self-attention mechanisms to capture global contextual dependencies, these models have achieved state-of-the-art performance in image classification tasks \cite{dosovitskiy2020image}.

One of the major challenges in applying DL to arsenic-induced skin lesion detection is the scarcity of suitable datasets. For skin cancer and other common dermatological conditions, several large-scale public repositories are available, such as the ISIC archive, DermNet, and HAM10000 \cite{ISIC2021, HAM10000}. However, these collections are composed primarily of dermoscopic images, which, although valuable for clinical practice, require specialized equipment and trained dermatologists to acquire. Such resources are rarely available in primary care or rural contexts, limiting their applicability in community-level screening. In contrast, arsenic-related skin lesion datasets are extremely rare and, to date, no large public repository exists. Only a handful of research efforts have explored image-based diagnosis of arsenicosis, highlighting a critical gap. This study seeks to address that gap by developing a DL framework for automated recognition of arsenic-induced skin lesions from non-dermoscopic images.

In this study, we focus on arsenic-induced skin lesion images captured using mobile devices, as these are far more practical in resource-constrained and rural settings compared to dermoscopic imaging. Merely distinguishing arsenic lesions from normal skin is not sufficient, since many arsenic-induced conditions closely resemble other dermatological disorders. Therefore, a robust diagnostic system must be evaluated against multiple clinically relevant skin conditions with overlapping visual features rather than a simple binary arsenic-versus-normal classification. To this end, we have curated a comprehensive dataset of non-dermoscopic, mobile-acquired skin images that includes arsenic-induced lesions alongside a broad spectrum of visually similar and common skin disorders. The dataset spans 20 diagnostic categories, providing a challenging and realistic benchmark for developing and evaluating AI-based diagnostic systems tailored to arsenic-affected populations.

We have utilized several DL architectures, including traditional CNNs and modern transformers. CNNs have long been the standard architecture, owing to their ability to hierarchically extract local features through convolutional kernels. They have demonstrated strong performance in a wide range of dermatological applications, particularly in tasks involving texture- and region-based differentiation of skin lesions \cite{choudhary2022skin}. However, CNNs are inherently limited in modeling long-range dependencies and global contextual relationships, which are often critical in distinguishing lesions with subtle variations. More recently, transformer-based architectures have been adapted to computer vision. These models leverage self-attention mechanisms to capture global dependencies across an image, enabling a more holistic representation of lesion patterns. Vision Transformers (ViTs) and their variants, such as Swin Transformer, have achieved state-of-the-art results in image classification benchmarks \cite{liu2021swin}. By comparing CNN- and transformer-based models on our curated dataset, we aim to evaluate their relative strengths and limitations in the classification of arsenic-induced skin lesions. To further assess generalizability, we also conducted external validation using images outside the curated dataset. The proposed framework showed promising results, demonstrating the potential applicability of the framework in real-world clinical settings.

A major barrier to adopting DL models in clinical practice is their lack of interpretability. Clinicians often require more than a prediction—they need insight into why a model arrived at a particular decision. To address this, our framework incorporates explainable AI (XAI) techniques, specifically Local Interpretable Model-agnostic Explanations (LIME) and Gradient-weighted Class Activation Mapping (Grad-CAM). LIME generates region-based perturbation maps that highlight the areas of an image most influential to the model’s classification \cite{Ribeiro2016} , while Grad-CAM produces heatmaps from the model’s internal feature activations to visualize class-discriminative regions  \cite{Selvaraju2017}. By combining these approaches, we provide both model-agnostic and model-specific interpretability, offering complementary insights into the decision-making process. These explanations are analyzed for both correctly classified and misclassified cases, enabling us to identify potential sources of error and iteratively refine the models. In the context of rural communities, where trained dermatologists are often unavailable, such interpretability tools are particularly valuable. Visual explanations can help frontline healthcare workers and non-specialists validate whether the model is focusing on clinically relevant lesion patterns, thereby improving trust, usability, and adoption. Moreover, XAI can assist in early triage, guiding referrals to specialized care when suspicious or high-risk patterns are detected.

To translate our research into a practical clinical tool, we developed and deployed a web-based application that serves the trained model via a lightweight inference API. Through a standard browser interface, users can upload skin lesion images, receive instant predictions with confidence scores, and view explainable overlays (LIME and Grad-CAM) that highlight regions influencing the decision. This real-time inference system can be used in arsenic-affected rural areas for preliminary screening and health triaging. 

\vspace{0.5cm}

The main contributions of this work can be summarized as follows:

\begin{itemize}
    \item Curated a dataset of skin lesion images, including arsenic-induced cases and 19 other conditions, to develop a DL framework on images captured by mobile phones.

    \item Benchmarked multiple DL architectures, including both CNNs and Transformer-based models, to assess their effectiveness in multiclass classification of skin lesions under data-limited conditions.

    \item Incorporated XAI methods, namely LIME and Grad-CAM, to interpret model predictions and highlight strengths and weaknesses in classification, particularly for misclassified arsenic cases.

    \item Developed a proof-of-concept web-based diagnostic tool that delivers predictions and visual explanations, demonstrating the feasibility of deploying the proposed framework in rural arsenic-affected communities.

    \item Validated generalization with external samples, showing that the framework extends reasonably well beyond the curated dataset.

    \item Investigated fine-tuning strategies and demonstrated that extensive unfreezing of pretrained backbones offers little benefit for small medical datasets, while frozen backbones with lightweight classifier heads remain more stable.

    \item Provided a detailed error analysis using confusion matrices and interpretability outputs, identifying common misclassification patterns and practical limitations of AI-based CAD systems in real-world settings.
\end{itemize}

To the best of our knowledge, this is the first fully integrated framework that focuses specifically on arsenic-affected skin lesion classification from mobile phone images, integrates explainability, and supports real-time deployment. All resources, including the dataset, trained model, web app, and code, are made publicly available to ensure reproducibility and facilitate future research in this direction.

The remainder of this paper is organized as follows. Section 2 reviews related work on arsenic detection and skin disease classification. Section 3 describes the data collection process.
The methods utilized in this study are discussed in detail in Section 4. Section 5 details the experimental setup and evaluation metrics. Section 6 reports and discusses the obtained results. Section 7 provides an in-depth analysis of the diagnostic framework, including model interpretability, error analysis, limitations, and the deployment strategy. Finally, Section 8 concludes the paper and outlines potential avenues for future research.

\section{Related Works}

In recent years, DL has emerged as a powerful tool in dermatological image analysis. With the release of several public skin lesion datasets, such as ISIC or HAM10000, research on automated skin lesion identification has progressed significantly \cite{hasan2023survey}. ISIC, which stands for  International Skin Imaging Collaboration, regularly organizes annual challenges, providing thousands of dermoscopic
images annotated by experts. Smaller datasets such as PH2, Derm7pt, and MED-NODE have also contributed to the development of DL-based CAD systems for skin lesion classification \cite{yan2025derm1m}. 

In a landmark study, Esteva et al. trained a deep CNN on over 120,000 skin images, demonstrating dermatologist-level accuracy in classifying malignant versus benign lesions \cite{esteva2017dermatologist}. This work established DL as a viable approach for computer-aided dermatological diagnosis and spurred extensive research into AI-powered skin lesion analysis.
Arora et al. conducted a comprehensive evaluation of fourteen state-of-the-art CNN architectures for multi-class skin lesion classification \cite{arora2023comparative}. The authors utilized the ISIC 2018 dataset, comprising approximately 10,154 dermoscopic images spanning seven diagnostic categories. Among them, DenseNet201 emerged as the top performer, achieving the highest classification accuracy of 82.5\%. Some researchers have attempted to merge multiple datasets to increase the diversity of the data. For example, Rafay and Hussain proposed EfficientSkinDis, an EfficientNet-based DL framework trained on images from 31 skin disease categories by combining samples from the ISIC archive and the Dermatology Atlas \cite{rafay2023efficientskindis}. Their approach leveraged transfer learning and data augmentation to overcome class imbalance and variability across datasets.

Most existing studies, along with the datasets on which they are trained, predominantly rely on dermoscopic images acquired under controlled lighting conditions with specialized equipment. While useful for clinical practice, such images do not adequately represent the variability of real-world mobile or primary-care photography. The limited availability of large-scale, high-quality smartphone-captured datasets poses a significant barrier to developing robust models and achieving practical deployment, especially in low-resource settings. Furthermore, these repositories are largely restricted to common dermatological conditions such as melanoma, with only a small number of images available for rare skin disorders. None of the widely used open datasets include images of arsenic-induced skin lesions, leaving this critical public health issue largely unexplored in existing CAD research.

Hsu et al. conducted one of the first non-invasive, image-based studies to predict arsenic exposure levels using hand and foot photographs from 2,497 subjects \cite{hsu2024rapid}. Employing ResNet-152 and ViTs, their models achieved an AUC of 0.81 for binary classification of arsenic versus normal cases. While this rapid, image-driven approach shows promise for clinical screening, the study did not account for visually similar non-arsenic skin conditions, raising the risk of misdiagnosis if deployed in real-world dermatological practice. Misdiagnosing another skin condition as arsenic could be harmful. Thus, it limits the model's applicability in real-world dermatological scenarios. Furthermore, the dataset used in this work remains confidential and inaccessible, limiting reproducibility and external validation.

ArsenicSkinImageBD was recently introduced as the only publicly available dataset of arsenic-induced skin lesions \cite{emu2024arsenicskinimagebd}. The dataset consists of mobile phone–captured photographs of common arsenicosis manifestations, collected from patients in arsenic-affected regions in Bangladesh. There are a total of 741 images of arsenic-affected people, and they have served as a primary source of arsenic-induced skin lesion data for this study.

Recently, Mehedi et al. introduced ArsenicNet, a fusion of Xception and Inception architectures, trained on the ArsenicSkinImageBD dataset \cite{mehedi2025arsenicnet}. Their model achieved around 97\% accuracy for a binary classification task (arsenic vs normal). In a similar study, Hossen et al. proposed a lightweight CNN architecture to distinguish arsenic-affected skin regions from normal ones \cite{hossen2025efficient}. Their proposed model achieved a testing accuracy of around 98\%.

Although these approaches demonstrate promising performance, their focus on binary classification (arsenic vs. normal) and reliance on relatively small datasets limit their clinical applicability and generalizability. Differentiating arsenic-affected skin from normal skin is often a relatively trivial task for DL models, as the visual differences are more pronounced than those between arsenic lesions and other dermatological conditions. Consequently, while prior binary classification studies report high accuracy, these results likely overestimate real-world performance. Distinguishing arsenicosis from visually similar non-arsenic skin diseases can be more challenging. Notably, these prior studies are unable to differentiate arsenic-induced lesions from other dermatological conditions with overlapping visual characteristics, which poses a significant risk of misdiagnosis in real-world practice. To date, there is no published work on multiclass models specifically tailored to arsenic-related skin lesion classification with broader disease comparisons.

In this study, we address this gap by developing a multiclass, mobile-acquired dataset and evaluating both CNN- and transformer-based DL models, complemented with XAI techniques and deployment strategies, to create a more robust and clinically relevant framework for arsenicosis detection.

\section{Data Curation}

For this study, we curated a custom dataset by compiling images from both publicly available repositories and online sources through web scraping to construct a comprehensive repository of skin diseases. We selected a total of 20 classes, with a particular focus on arsenic-induced skin lesions as well as a set of visually similar and common dermatological conditions such as keratosis, carcinomas, pox, measles, and some other inflammatory and infectious skin disorders. A normal skin class was also included as a baseline reference. The dataset contains over 11000 images and includes only non-dermoscopic images, captured using smartphone cameras. Image quality varied due to different smartphone cameras and lighting conditions. By incorporating these confounding conditions, we aimed to reflect real-world diagnostic complexity. This enables our models not only to distinguish arsenicosis from normal skin but also to identify other co-occurring skin conditions that a patient may be suffering from, thereby increasing the clinical relevance of the framework. 

\begin{figure}[htbp]
    \centering
    \includegraphics[width=0.9\textwidth]{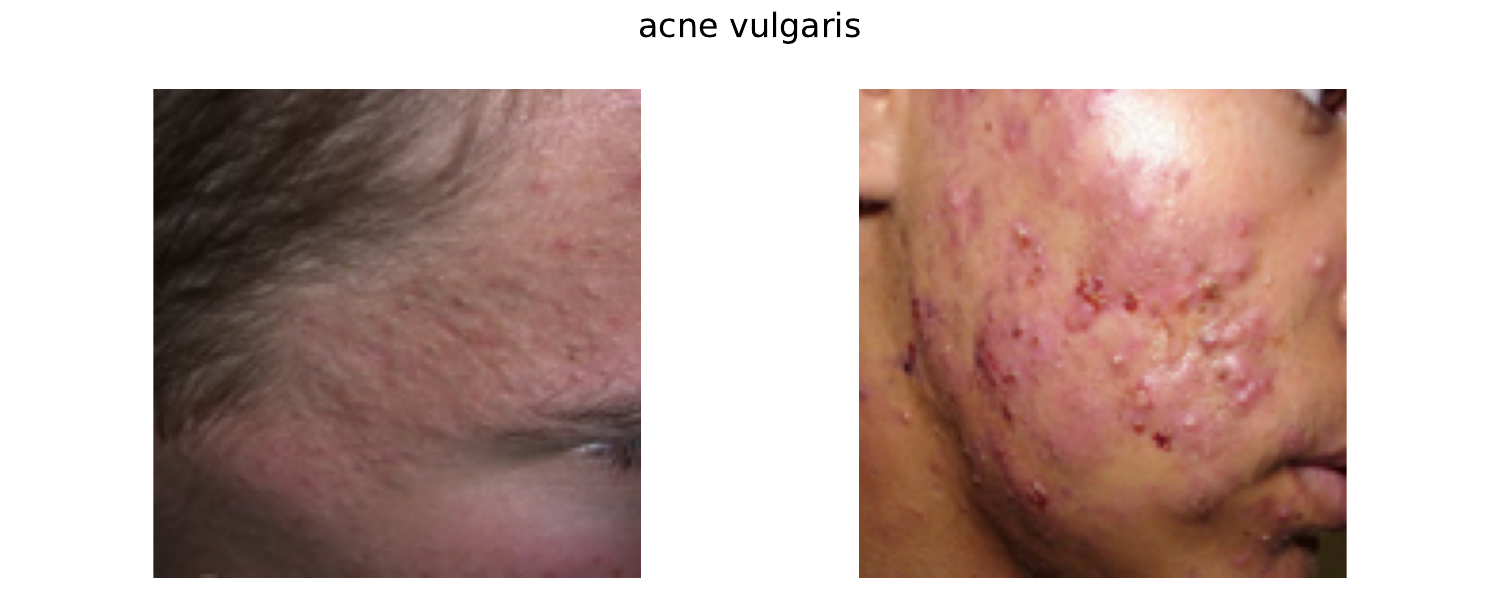}\\[0.5em]
    \includegraphics[width=0.9\textwidth]{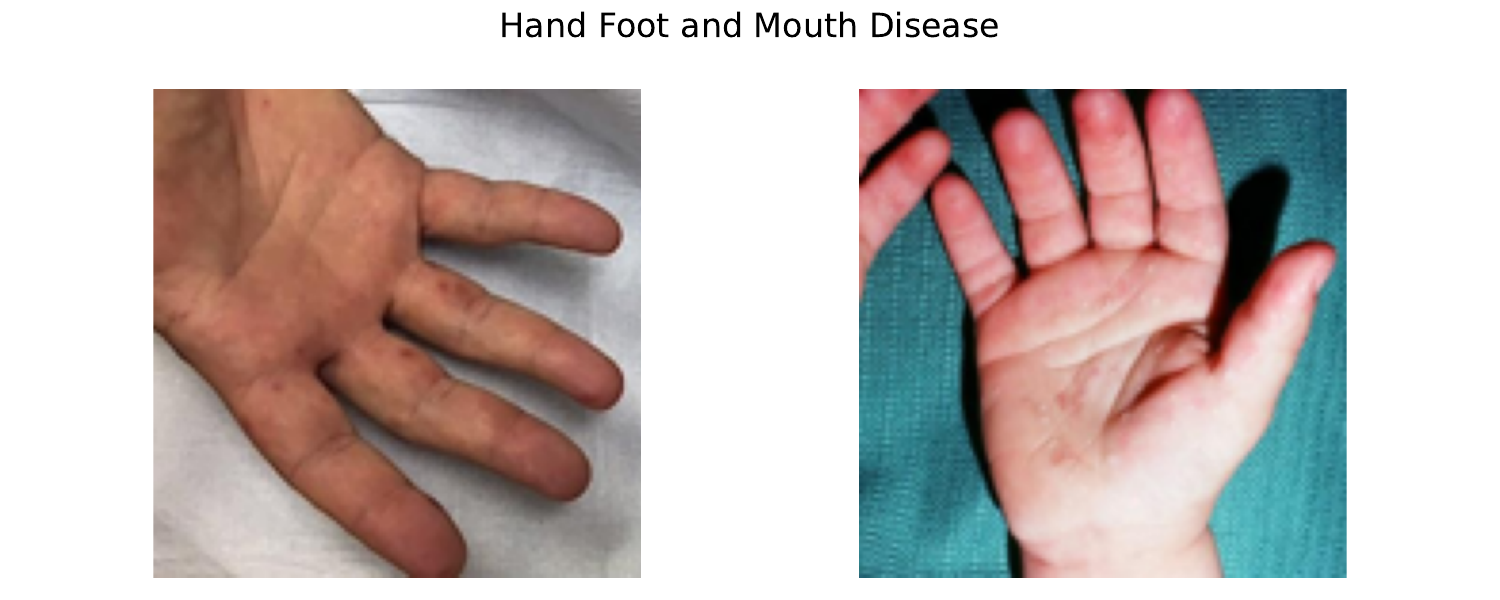}
    \caption{Some sample images from the curated dataset.}
    \label{fig:dataset_samples}
\end{figure}

A summary of the curated dataset is provided in Table \ref{tab:dataset_distribution}. The public repositories that were used to collect certain samples are listed in Table \ref{tab:public_sources}. In addition to these repositories, we also collected several hundred supplementary images from online sources and publicly available web platforms, which were carefully reviewed to ensure relevance and quality. Some sample images from the dataset are shown in Figure \ref{fig:dataset_samples}. In future work, we plan to expand this dataset by incorporating additional skin conditions and collecting more geographically diverse samples. This will further improve the robustness and generalizability of the proposed framework.

\begin{table}[htbp]
\centering
\caption{Distribution of images across the 20 classes in the curated dataset.}
\label{tab:dataset_distribution}
\begin{tabular}{l r l r}
\hline
\textbf{Class} & \textbf{No.} & \textbf{Class} & \textbf{No.} \\
\hline
Arsenic & 819 & Chickenpox & 482 \\
Actinic Keratosis & 951 & Cowpox & 330 \\
Basal Cell Carcinoma & 1599 & Hand Foot and Mouth Disease & 805 \\
Squamous Cell Carcinoma & 730 & Measles & 366 \\
Melanoma & 343 & Monkeypox & 1699 \\
Nevus & 244 & Tinea Corporis & 176 \\
Seborrheic Keratosis & 321 & Scabies & 314 \\
Acne Vulgaris & 393 & Lichen Planus & 474 \\
Seborrheic Dermatitis & 181 & Pityriasis Versicolor & 148 \\
Vitiligo & 205 & Normal & 1409 \\
\hline
\textbf{Total} & \textbf{11,489} & & \\
\hline
\end{tabular}
\end{table}

\begin{table}[htbp]
\centering
\caption{Public repositories used for dataset curation.}
\label{tab:public_sources}
%\begin{tabular}{l l l}
\begin{tabularx}{\textwidth}{X l X}
\hline
\textbf{Repository} & \textbf{Reference} & \textbf{Categories Available} \\
\hline
ArsenicSkinImageBD & \href{https://data.mendeley.com/datasets/x4hgnjj5gv/2}{Mendeley Data} & Arsenic-induced skin lesions, Normal \\
MSID (Monkeypox Skin Images Dataset) & \href{https://data.mendeley.com/datasets/r9bfpnvyxr/6}{Mendeley Data} & Monkeypox, Chickenpox, Measles, Normal \\
MSLD2.0 (Mpox Skin Lesion Dataset v2.0) & \href{https://www.kaggle.com/datasets/joydippaul/mpox-skin-lesion-dataset-version-20-msld-v20}{Kaggle} & Monkeypox, Chickenpox, Cowpox, Measles, Normal \\
PAD-UFES-20 & \href{https://data.mendeley.com/datasets/zr7vgbcyr2/1}{Mendeley Data} & Basal cell carcinoma, \\
 & & Squamous cell carcinoma, Actinic keratosis, Melanoma, Nevus\\
Dermatology Atlas & \href{https://www.atlasdermatologico.com.br/}{Online} & Multiple dermatological conditions (e.g., acne, eczema, vitiligo, etc.) \\
\hline
\end{tabularx}
\end{table}

\section{Methodology}

\subsection{Deep Learning Architectures}

In this study, a broad set of DL architectures, including both
CNN-based and transformer-based models were evaluated to
establish a robust baseline and identify the most suitable model
for the skin lesion classification task. CNNs have been the backbone of medical image analysis for over a decade, effectively capturing local patterns such as texture, edges, and lesion boundaries that are critical for skin disease classification \cite{young2019deep}. In contrast, transformer-based architectures employ self-attention mechanisms to model long-range dependencies and global context across the entire image, enabling them to capture subtle structural variations that CNNs may overlook. The DL architectures utilized in this study are listed in Table \ref{tab:models}. These models are briefly described below.

\begin{table}[htbp]
\centering
\caption{DL architectures evaluated for arsenic-induced skin disease classification.}
\label{tab:models}
\begin{tabular}{|p{0.28\textwidth}|p{0.62\textwidth}|}
\hline
\textbf{Category} & \textbf{Models} \\
\hline
CNN-based & VGG16 \cite{simonyan2014very}\\
& ResNet50 \cite{he2016deep}\\
& InceptionV3 \cite{szegedy2016rethinking}\\
& Xception \cite{chollet2017xception}\\
& EfficientNetB0 \cite{tan2019efficientnet}\\
& MobileNetV2 \cite{howard2019searching}\\
\hline
Transformer-based & Vision Transformer (ViT-B/16) \cite{dosovitskiy2020image}\\
& Swin Transformer \cite{liu2021swin}\\
%& DeiT (Data-efficient Image Transformer) \cite{touvron2021training} \\
\hline
Transformer-inspired CNN & ConvNeXt \cite{liu2022convnet}\\
\hline
\end{tabular}
\end{table}

\subsubsection{CNN-based Architectures}

\textbf{VGG16}: A deep CNN with 16 layers that uses small 3×3 filters and sequential convolution-pooling blocks. It is simple yet effective in capturing hierarchical visual features. Although VGG16 is widely used as a baseline due to its depth and simplicity, its large parameter count makes it computationally expensive and less suitable for fine-grained lesion classification compared to more modern architectures.

\textbf{ResNet50}: Introduces residual connections (skip connections) to allow very deep networks to train effectively by mitigating the vanishing gradient problem. While ResNet50 provides good generalization, it is computationally heavier compared to other models.

\textbf{InceptionV3}: Employs “Inception modules” with multiple filter sizes in parallel to capture both local and global features efficiently, reducing parameter count. For example, a $3 \times 3$ convolution is factorized into sequential $1 \times 3$ and $3 \times 1$ convolutions, which reduces redundancy and computational cost while retaining representational power. The use of multi-scale kernels within each module enables effective capture of both coarse and fine lesion features, making InceptionV3 more efficient and accurate than VGG16 for complex medical imaging tasks.

\textbf{Xception}: An extension of Inception that can be viewed as an “extreme” version of Inception, where standard convolutions are replaced entirely by depthwise separable convolutions. This decomposition improves parameter efficiency while maintaining high representational capacity.

\textbf{EfficientNetB0}: Uses a compound scaling method to balance network depth, width, and resolution, achieving high accuracy with fewer parameters. Its lightweight design provides
a good accuracy-efficiency tradeoff.

\textbf{MobileNetV2}: Lightweight CNN optimized for mobile and embedded devices; uses depthwise separable convolutions and inverted residuals to reduce computation. This architecture provides fast inference but achieves lower accuracy compared to deeper models.

\subsubsection{Transformer-based Architectures}

\textbf{Vision Transformer}: It is the first pure transformer applied to vision tasks. Splits images into fixed-size patches and processes them as token sequences using self-attention. ViT captures long-range dependencies effectively but usually requires very large datasets for successful training.

%\textbf{Data-efficient Image Transformer}: DeiT alleviates ViT’s dependence on massive datasets by employing \textit{knowledge distillation} from a CNN teacher (e.g., ResNet). This allows it to achieve competitive results without requiring huge pretraining corpora.

%$\mathcal{O}(N^2)$ (where $N$ is the number of image patches)

\textbf{Swin Transformer}: A hierarchical transformer with shifted windows that computes attention locally and globally. Unlike ViT, which applies global self-attention with quadratic complexity, the Swin Transformer restricts attention to non-overlapping windows of size $M \times M$, reducing overall complexity. This design enables it to capture both local patterns (e.g., surface roughness, pigmentation irregularities) and global lesion structures (e.g., lesion boundaries relative to surrounding skin). Such multi-scale attention is particularly advantageous in medical imaging, where both local details and contextual cues are critical. 

%\begin{equation}
%\mathcal{O}(M^2 \cdot N)
%\end{equation}
%where $M \ll N$, making it highly scalable for high-resolution medical images. 

\subsubsection{Transformer-inspired CNN}

\textbf{ConvNeXt}: A modernized CNN architecture that integrates design principles from transformers (e.g., large kernels, layer normalization). It combines CNN efficiency with transformer-like performance on image classification tasks.

\subsection{Explainable AI (XAI)}

A major challenge in deploying DL models in healthcare is their “black-box” nature, which limits transparency and clinical trust. XAI techniques aim to bridge this gap by providing human-interpretable insights into model predictions \cite{newaz2024ml}. In this study, we have utilized two XAI methods: LIME and Grad-CAM. These methods generate visual explanations that highlight the image regions most influential in decision-making. These explanations not only enhance clinical confidence but also support error analysis, model refinement, and safer deployment in real-world screening scenarios.

\subsubsection{LIME}

LIME is a widely used technique for interpreting predictions from any machine learning classifier. It is a model-agnostic XAI technique that explains predictions by approximating the complex model locally with a simpler, interpretable surrogate (e.g., linear regression). For image classification, LIME perturbs superpixel regions of the input and observes changes in prediction probabilities to identify which parts of the image contribute most to the decision ~\cite{Ribeiro2016}. This allows clinicians to visualize whether the model is focusing on medically relevant lesion regions or being influenced by irrelevant artifacts.

\subsubsection{Grad-CAM}

Grad-CAM is a model-specific explanation method that produces heatmaps highlighting the spatial regions most important for a particular prediction. It works by backpropagating the gradients of the target class into the final convolutional layers, using these to generate class-discriminative localization maps~\cite{Selvaraju2017}. Grad-CAM is especially useful in medical imaging because it leverages the model’s internal feature representations, enabling clinicians to assess whether the decision is grounded in pathology-relevant features \cite{suara2023grad}.

\subsection{Model Deployment}

For practical applicability, the trained models were integrated into a deployment framework designed to support real-time use in clinical and community settings. Deployment can be carried out through two complementary pathways. In this study, a web-based application was developed to allow users to upload skin lesion images via a browser interface, obtain rapid predictions with confidence scores, and visualize explanation overlays (Grad-CAM and LIME). This server-based approach simplifies model updates, ensures consistency across users, and supports large-scale accessibility. In addition to this, we also propose the design of a lightweight mobile application for on-device inference, which would be particularly valuable in rural arsenic-affected regions where internet connectivity and access to dermatologists are limited. The development of this mobile application is planned for future work.

\section{Experimental Evaluation}

\subsection{Model Utilization}

In this study, we experimented with 10 DL architectures. For each model, we employed transfer learning with ImageNet-pretrained weights. These weights are derived from training on over one million natural images. They provide generic low- and mid-level feature representations (e.g., edges, textures, and shapes). Leveraging such pretrained features is particularly useful in our task, where the number of samples per class is relatively small. Starting from pretrained weights, rather than random initialization, enables the models to benefit from strong transferable features and improves convergence.

The convolutional and transformer backbones of the DL models were initially frozen, and a custom classification head was added. The earlier layers perform the feature extraction task, and the final layers (head) perform classification. This head consisted of a global average pooling (GAP) layer, followed by one or two fully connected layers (128/256 neurons) with ReLU activation, a batch normalization layer, dropout for regularization, and a final softmax output. During training, the backbone weights remained unchanged. In each epoch, the classification head (dense layers) learns and updates its weights according to the multiclass classification task at hand to learn task-specific features. 

\begin{figure}[htbp]
    \centering
    \includegraphics[width=0.9\textwidth]{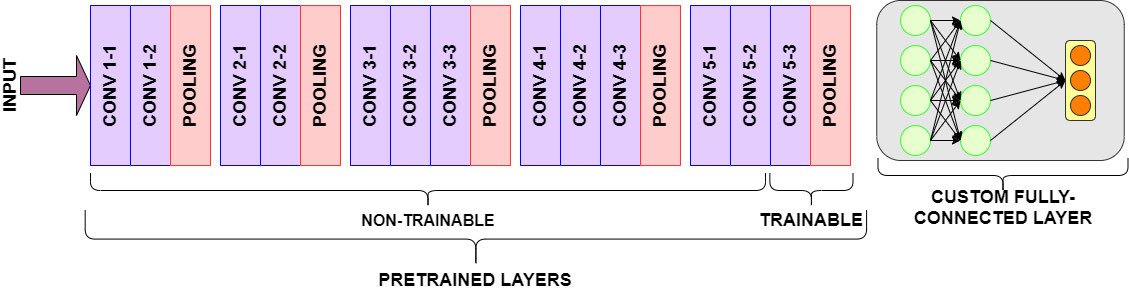}
    \caption{Illustration of the fine-tuning strategy employed in this study. Adapted from \cite{learndatasci_transferlearning}.
    }
    \label{fig:transfer_learning}
\end{figure}

In addition to this, we also experimented with partial fine-tuning by unfreezing a few backbone layers. This is illustrated in Figure \ref{fig:transfer_learning}. Fine-tuning allows convolutional filters to adapt to new domain-specific features, which can be beneficial when sufficient data is available. However, in our case, the dataset was relatively small and comprised numerous categories with subtle visual differences. As a result, fine-tuning generally led to overfitting and reduced performance. The limited number of samples per class and the high similarity between arsenic and non-arsenic conditions likely contributed to this. Ultimately, training with frozen backbones and a custom classification head produced the most stable and generalizable results for the arsenic dataset.

\subsection{Experimental Framework}

To ensure a rigorous evaluation, we designed a structured experimental framework covering dataset preparation, preprocessing, training, and evaluation. The dataset was divided into training (60\%), validation (20\%), and test (20\%) subsets. Training set images were used for model learning, while the validation set was used to monitor overfitting during training, tuning hyperparameters, and selecting the best model checkpoint. The final performance of the trained model was evaluated on the test set. The original images varied in resolution. Different models require different input image sizes. All the images were therefore resized according to the model's needs before training. Pixel intensities were normalized using the ImageNet mean and standard deviation to align with the pretrained model weights.

Details about different training parameter settings are reported in Table \ref{tab:exp_setup}. The framework was implemented in TensorFlow/Keras and PyTorch, and experiments were run on cloud-based environments (Google Colab and Kaggle).

\begin{table}[htbp]
\centering
\caption{Experimental setup and training configuration.}
\begin{tabular}{ll}
\hline
\textbf{Parameter} & \textbf{Value} \\
\hline
Dataset Split      & 60\% Train, 20\% Validation, 20\% Test \\
Loss Function      & Categorical Cross-Entropy \\
Image Sizes        & 224$\times$224 (most CNNs and transformers), \\
& 299$\times$299 (InceptionV3, Xception) \\
Optimizer          & Adam \\
Learning Rate      & $1 \times 10^{-4}$ \\
Batch Size         & 32 \\
Epochs             & Up to 50 (with early stopping) \\
%Normalization      & ImageNet mean and standard deviation \\
Regularization     & Dropout (0.3 or 0.4), Batch Normalization \\
Callbacks          & EarlyStopping, ReduceLROnPlateau, ModelCheckpoint \\
Hardware           & Kaggle GPU (NVIDIA Tesla T4) \\
\hline
\end{tabular}
\label{tab:exp_setup}
\end{table}

\subsection{Evaluation Metrics}

To thoroughly assess the performance of the proposed model across all 20 skin disease classes, we utilized five different evaluation metrics: Accuracy, Precision, Recall, F1-score, and Matthews Correlation Coefficient (MCC). Since the dataset is imbalanced across classes, weighted averages of Precision, Recall, and F1-score were reported to account for class support, ensuring that larger classes did not disproportionately dominate the performance evaluation.

Accuracy measures the proportion of correctly classified samples among all predictions and provides a general sense of the
model’s overall effectiveness. Precision is the fraction of correctly identified positive samples among all samples predicted as positive, reflecting the model’s ability to avoid false positives. High precision ensures that the model is not over-predicting any
particular class, which is crucial when some skin diseases are
rare. Recall (Sensitivity) is the fraction of correctly identified positive samples among all actual positives, reflecting the model’s ability to capture true cases and avoid false negatives. High recall ensures that instances of each class, including less frequent or subtle lesions, are effectively detected. F1-score is the harmonic mean of Precision and Recall, offering a balanced evaluation of performance when both false positives and false negatives are important. MCC is widely regarded as one of the most reliable metrics for imbalanced data, as it considers all entries of the confusion matrix and provides a balanced evaluation even when class distributions are skewed \cite{newaz2024icost, halimu2019empirical}.

\paragraph{Accuracy}
\begin{equation}
\text{Accuracy} = \frac{\sum_{i=1}^{K} TP_i}{N}
\end{equation}
where $TP_i$ denotes the number of correctly classified samples for class $i$, $K$ is the total number of classes, and $N$ is the total number of samples.

\paragraph{Precision}
\begin{equation}
\text{Precision}_i = \frac{TP_i}{TP_i + FP_i}
\end{equation}

\paragraph{Recall}
\begin{equation}
\text{Recall}_i = \frac{TP_i}{TP_i + FN_i}
\end{equation}

\paragraph{F1-score}
\begin{equation}
\text{F1}_i = 2 \cdot \frac{\text{Precision}_i \cdot \text{Recall}_i}{\text{Precision}_i + \text{Recall}_i}
\end{equation}

\paragraph{MCC}
\begin{equation}
\text{MCC} = \frac{ TP \times TN - FP \times FN }{\sqrt{(TP+FP)(TP+FN)(TN+FP)(TN+FN)}}
\end{equation}
where $TP$, $TN$, $FP$, and $FN$ are computed over all classes.

\paragraph{Weighted Averaging}
Precision, Recall, and F1-score are reported as weighted averages:
\begin{equation}
\text{Metric}_{weighted} = \frac{\sum_{i=1}^{K} n_i \cdot \text{Metric}_i}{\sum_{i=1}^{K} n_i}
\end{equation}
where $n_i$ is the number of samples in class $i$.

\section{Results}

\subsection{Performance Measures}

In this section, we report the performance metrics obtained from all DL models. The evaluation was conducted on the held-out test set, which was kept completely separate from the training and validation data to ensure an unbiased assessment. A comprehensive summary of the results is provided in Table~\ref{tab:dl_model_performance}, which lists the performance measures for each model across all evaluation metrics.

\begin{table} [htbp]
\centering
\caption{Performance metrics of different DL models on the test set.}
\label{tab:dl_model_performance}
\begin{tabular}{lrrrrr}
\hline
         Model &  Accuracy &  Recall &  Precision &  F1 Score &  MCC \\
\hline

           VGG-16 &      0.74 &    0.74 &       0.73 &      0.72 & 0.71 \\
     Inception-V3 &      0.76 &    0.76 &       0.76 &      0.76 & 0.74 \\     
     ResNet-50 & 0.76 & 0.68 & 0.75 & 0.69 & 0.74 \\
     MobileNet-V2 &      0.78 &    0.78 &       0.78 &      0.77 & 0.76 \\
   
EfficientNet-B0 &      0.79 &    0.79 &       0.79 &      0.78 & 0.77 \\
Xception &      0.78 &    0.78 &       0.77 &      0.77 & 0.76 \\
ViT &      0.84 &    0.84 &       \textbf{0.86} &      0.84 & 0.83 \\
Swin &      \textbf{0.86} &    \textbf{0.86} &       \textbf{0.86} &      \textbf{0.86} & \textbf{0.85} \\
      ConvNeXt &      0.77 &    0.77 &       0.77 &      0.76 & 0.75 \\
\hline
\end{tabular}
\end{table}

Among the CNN-based architectures, Xception, EfficientNet-B0, and MobileNetV2 achieved comparable results, with EfficientNet-B0 attaining the highest MCC (0.77). In contrast, VGG16 and ResNet-50 performed worse. ConvNeXt achieved moderate performance but did not surpass the stronger CNN baselines. Transformer-based models, however, delivered the best overall results, with both the ViT and the Swin Transformer clearly outperforming all CNN variants. In particular, the Swin Transformer obtained the highest accuracy (0.86) and MCC (0.85), underscoring the advantage of self-attention mechanisms in modeling global dependencies and subtle discriminative patterns in the dataset.

\subsection{Class-Wise Performance}

\begin{figure}[htbp]
    \centering
    % ----------- First Image -----------
    \begin{subfigure}[b]{0.95\textwidth}
        \centering
        \includegraphics[height=0.4\textheight]{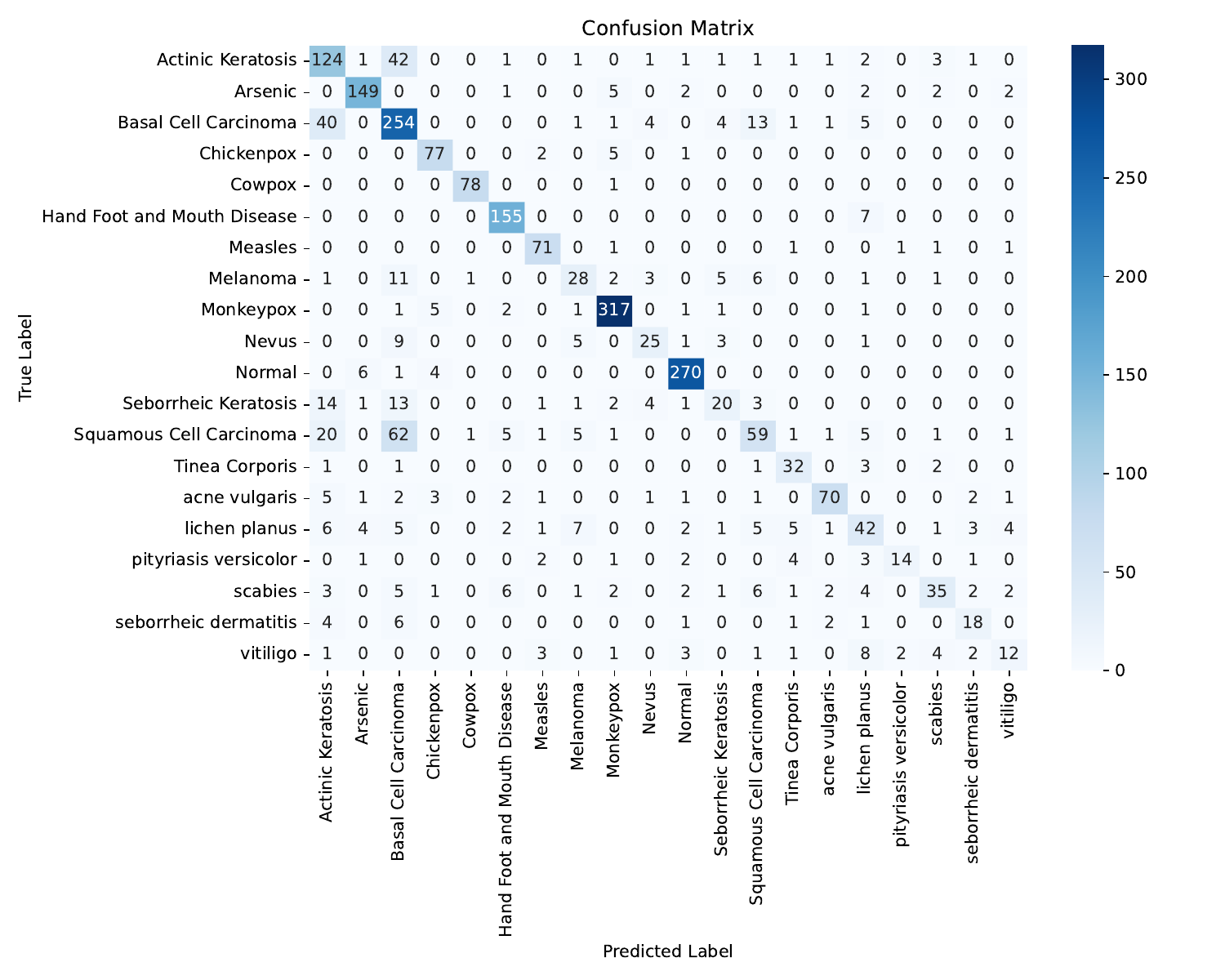}
        \caption{ConvNeXt confusion matrix}
        \label{fig:conf_a}
    \end{subfigure}
    \vskip\baselineskip
    % ----------- Second Image -----------
    \begin{subfigure}[b]{0.95\textwidth}
        \centering
        \includegraphics[height=0.4\textheight]{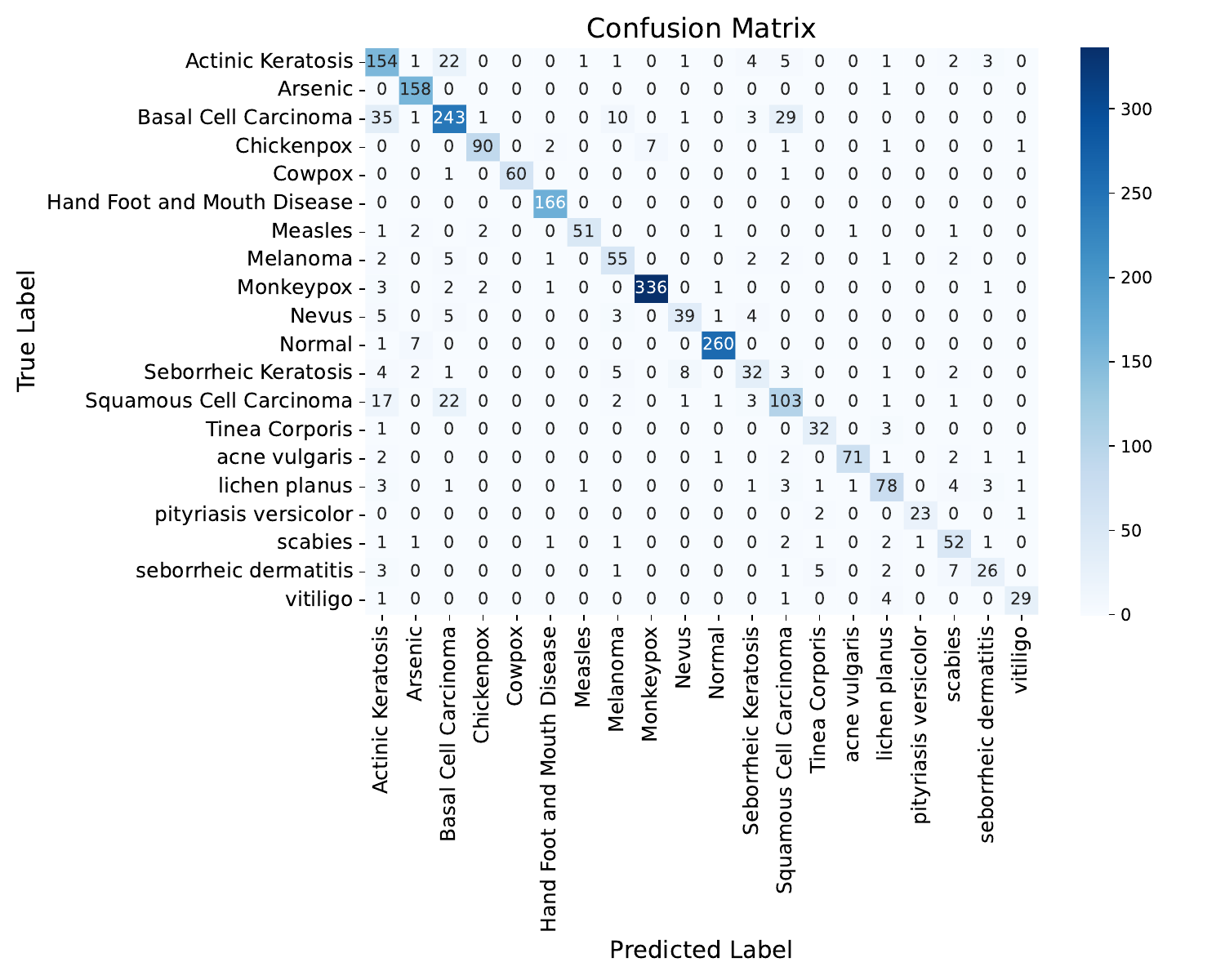}
        \caption{Swin Transformer confusion matrix}
        \label{fig:conf_b}
    \end{subfigure}
    
    \caption{Confusion matrices obtained from two different DL models.}
    \label{fig:conf_mat_examples}
\end{figure}

In this section, we examine the class-wise performance of the models using confusion matrices. Figure~\ref{fig:conf_mat_examples} presents two representative confusion matrices obtained from ConvNeXt and Swin Transformer. As illustrated in Figure~\ref{fig:conf_a}, the ConvNeXt model correctly classified 149 arsenic test samples while misclassifying approximately 10. The most frequent errors occurred in the categories BCC and SCC, which were often predicted as actinic keratosis (AK) or misclassified with one another. This confusion is clinically plausible, as BCC and SCC can exhibit overlapping visual characteristics such as erythematous patches, keratotic surfaces, and irregular pigmentation, particularly in early presentations \cite{ryu2018features}. These similarities, combined with the variability of mobile-captured images, make them especially difficult to distinguish in automated classification (discussed in more detail in Section 7.2).

Similar patterns of misclassification were also observed in the Swin Transformer model (Figure~\ref{fig:conf_b}); however, the overall number of errors was notably lower. The model misclassified only a single arsenic sample, demonstrating greater robustness in arsenic detection compared to the ConvNeXt model. This particular misclassified case, along with its underlying explanation, is further analyzed using LIME in section 7.1.1.

\subsection{Training Dynamics}

To further analyze the learning dynamics of the models, we examined the training and validation accuracy–loss curves for representative CNN- and Transformer-based architectures. Three examples are presented here, while the complete set of curves is provided in the GitHub repository. 

\begin{figure}[htbp]
    \centering
        % ----------- EfficientNetB0 -----------
    \includegraphics[width=0.9\textwidth]{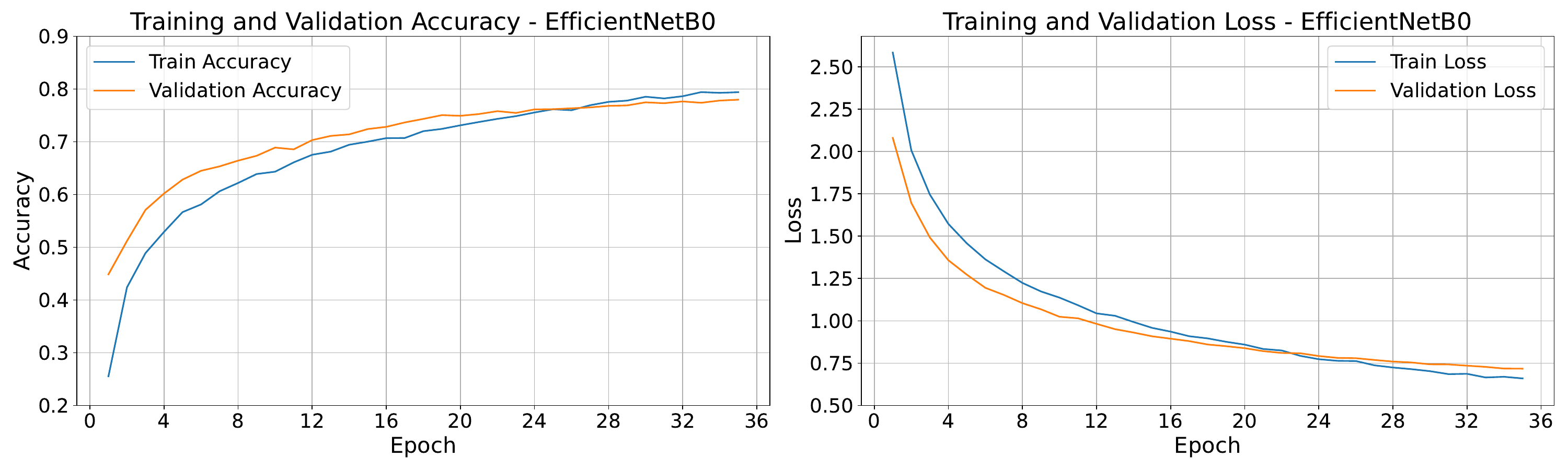}
    \caption{Training and validation curves for EfficientNetB0.}
    \label{fig:effnet_curves}

    \vskip\baselineskip
    % ----------- Xception -----------
    \includegraphics[width=0.9\textwidth]{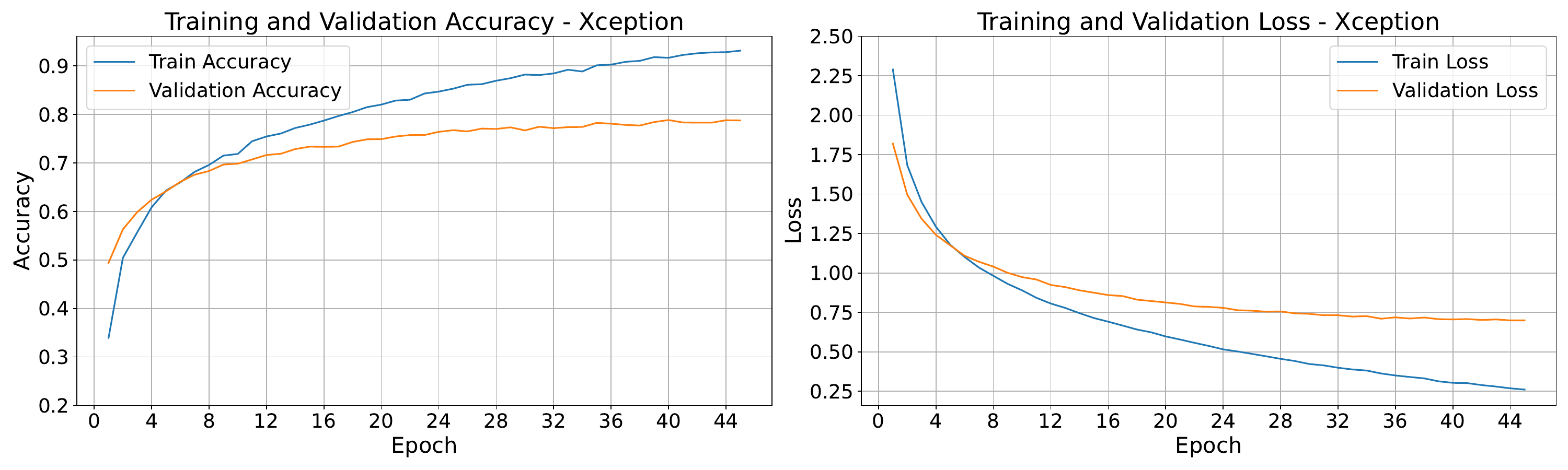}
    \caption{Training and validation curves for Xception.}
    \label{fig:xception_curves}
    
    \vskip\baselineskip % add space between figures
    
    % ----------- Swin Transformer -----------
    \includegraphics[width=0.9\textwidth]{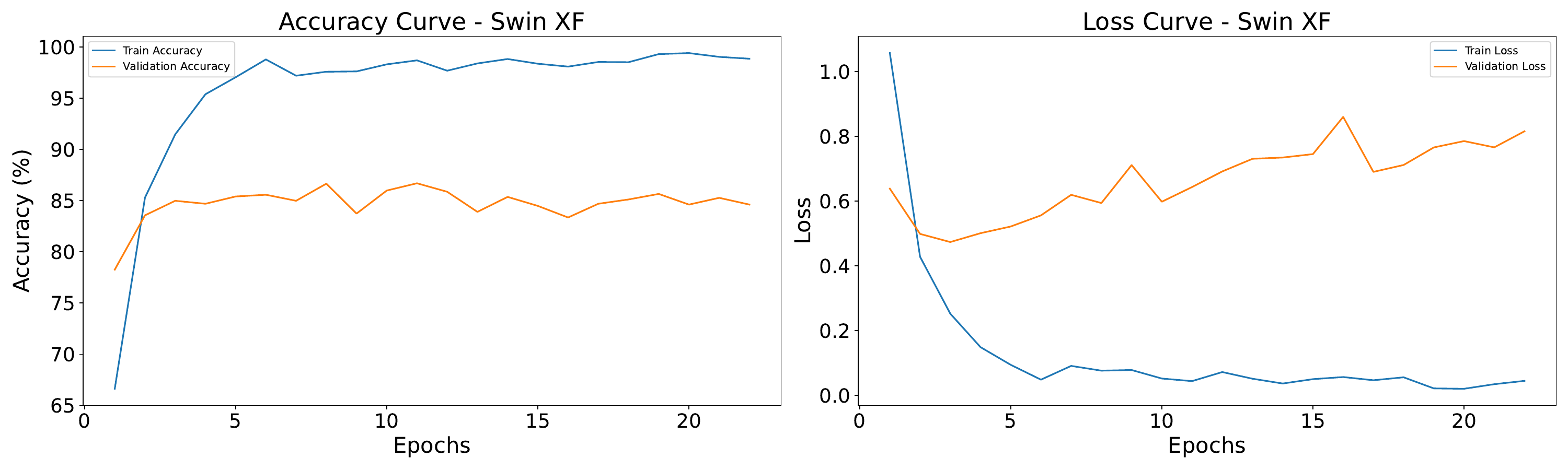}
    \caption{Training and validation curves for the Swin Transformer.}
    \label{fig:swin_curves}

\end{figure}

Some of the CNN-based models, such as EfficientNetB0, demonstrated steady convergence, with both training and validation accuracy improving gradually over the epochs (Figure~\ref{fig:effnet_curves}). During the initial 25 epochs, the validation loss remained lower than the training loss, indicating effective learning and good generalization. Beyond this point, the validation loss rose slightly above the training loss, though the gap between them remained small. This suggests limited to no overfitting, even as performance plateaued around 80\% validation accuracy. The ConvNeXt model exhibited a nearly identical learning pattern, with stable convergence and only a modest gap between training and validation performance.  

In contrast, the Xception model (Figure~\ref{fig:xception_curves}) achieved higher training accuracy but exhibited a widening gap between training and validation curves after 20 epochs, indicating mild overfitting despite a continuing decline in validation loss. This highlights the trade-off between model capacity and generalization in CNNs trained on relatively small medical datasets. Other CNNs, such as MobileNetV2 or VGG16, quickly overfitted within a few epochs.

The Swin Transformer (Figure~\ref{fig:swin_curves}) showed a markedly different behavior, converging rapidly within the first 10 epochs. Training accuracy approached 98\%, while validation accuracy plateaued around 85\%. Shortly thereafter, validation loss began to rise, a clear signal of overfitting. This rapid convergence is attributed to the self-attention mechanism, which allows the Transformer to efficiently capture long-range contextual dependencies. However, the limited dataset size constrained its ability to generalize fully, necessitating the use of early stopping to avoid severe overfitting. Nevertheless, a model checkpoint callback was used to automatically save and restore the weights corresponding to the best validation performance, ensuring that the final reported results were obtained before the onset of overfitting. Despite these challenges, the Swin Transformer still delivered the best performance, achieving 86\% accuracy on the test set.

\section{Discussion}

\subsection{Model Interpretation}

\subsubsection{LIME-based interpretability}

To enhance the interpretability of the proposed framework, we employed LIME to generate region-based explanations for model predictions. The resulting visualizations from LIME highlight the regions of the image that most strongly influenced the model’s classification. This enables verification of whether the model is focusing on clinically relevant areas, which is particularly important in arsenic-affected rural contexts where expert dermatologists are not readily available.

\begin{figure}[htbp]
    \centering
    % ----------- First Image -----------
    \begin{subfigure}[b]{0.9\textwidth}
        \centering
        \includegraphics[width=\linewidth]{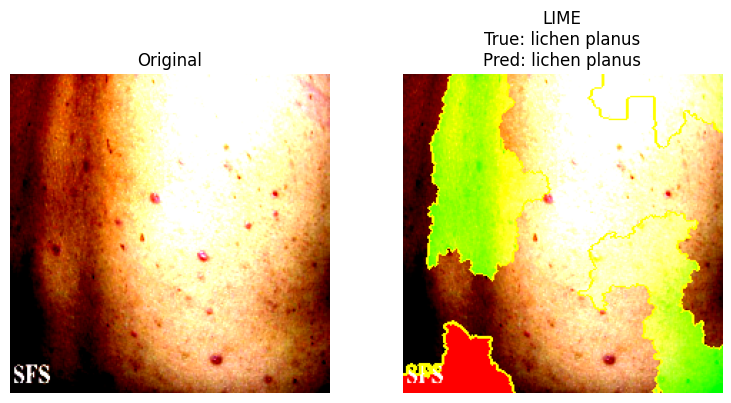}
        \caption{Correct classification.}
        \label{fig:lime_correct}
    \end{subfigure}
    \hfill
    % ----------- Second Image -----------
    \begin{subfigure}[b]{0.9\textwidth}
        \centering
        \includegraphics[width=\linewidth]{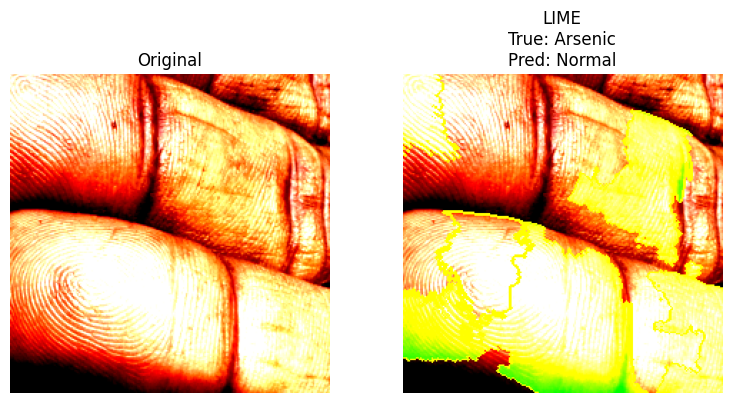}
        \caption{Misclassification.}
        \label{fig:lime_misclassified}
    \end{subfigure}
    
    \caption{Examples of LIME-based explanations. 
    (a) A correctly classified case where highlighted regions correspond to lesion-relevant features. 
    (b) A misclassified case where the model focused on normal-like skin areas due to poor image capture and lack of visible lesions.}
    \label{fig:lime_examples}
\end{figure}

Figure~\ref{fig:lime_correct} illustrates an example of LIME-based explainability for a random test case. The original image is shown alongside its LIME visualization. In this case, the model correctly classified the lesion as lichen planus. The highlighted superpixels indicate the regions that influenced the model’s prediction: green superpixels represent regions that positively influenced the model’s decision toward the predicted class, whereas red superpixels indicate regions that negatively influenced the prediction (in this case, the dark border area in the image), pulling the decision away from the target class. Here, LIME highlights the small dark papular lesions, which are consistent with the clinical presentation of lichen planus. This suggests that the model is attending to relevant dermatological features rather than background noise.

Figure~\ref{fig:lime_misclassified} illustrates a false-negative test case explained with LIME. Although the ground-truth label is arsenic, the model predicted it to be normal. The LIME attribution shows green superpixels concentrated on smooth palmar ridges and well-lit, non-lesional areas. Regions where punctate hyperkeratosis or melanosis would be expected are absent from the field of view in this poorly captured image. Given the subtlety of early arsenicosis and the high visual similarity to normal palmar texture, the network learned to trust the dominant “normal-like” cues in this image, resulting in a normal prediction. 

In short, the error is driven less by the classifier and more by acquisition quality and framing: the photograph does not clearly present lesion-bearing regions, so the model’s most influential evidence comes from non-lesional skin. This example also highlights a key weakness in the acquired data—not all images are of sufficient quality or capture the lesion context accurately. Such limitations are common in mobile-captured images obtained under real-world conditions, where variations in lighting, focus, framing, and lesion visibility are unavoidable. These factors can reduce the model’s ability to attend to clinically relevant features, ultimately leading to misclassifications.

\subsubsection{Grad-CAM based interpretability}

To further enhance interpretability, we employed Grad-CAM. Grad-CAM generates class-specific heatmaps by backpropagating gradients from the final classification layer to the last convolutional layer. The resulting activation maps highlight spatial regions that most strongly influenced the model’s prediction, thereby providing a visual explanation of the decision-making process.

\begin{figure}[htbp]
    \centering
    % ----------- First Image -----------
    \begin{subfigure}[b]{0.9\textwidth}
        \centering
        \includegraphics[width=\linewidth]{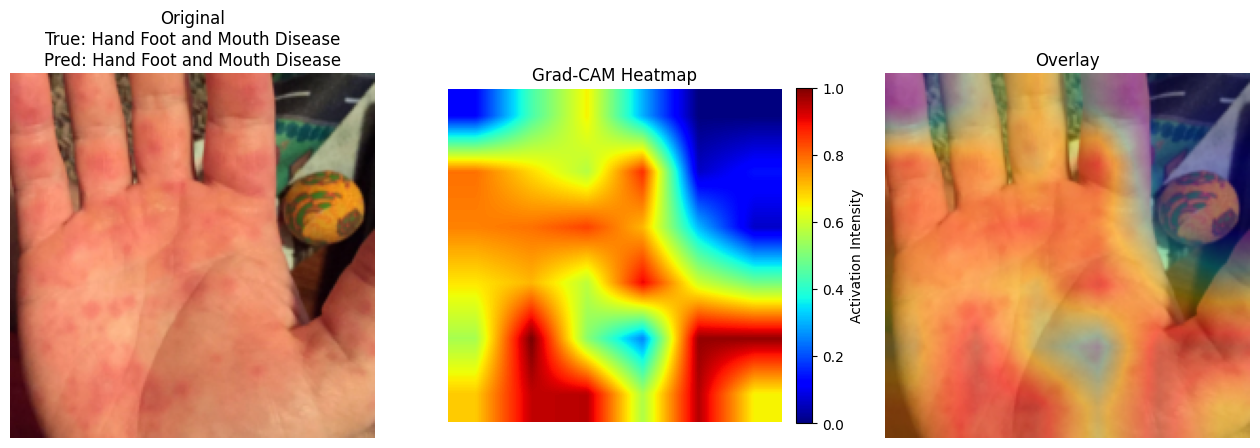}
        \caption{Hand, Foot and Mouth Disease.}
        \label{fig: Hand, Foot and Mouth Disease}
    \end{subfigure}
    \hfill
    % ----------- Second Image -----------
    \begin{subfigure}[b]{0.9\textwidth}
        \centering
        \includegraphics[width=\linewidth]{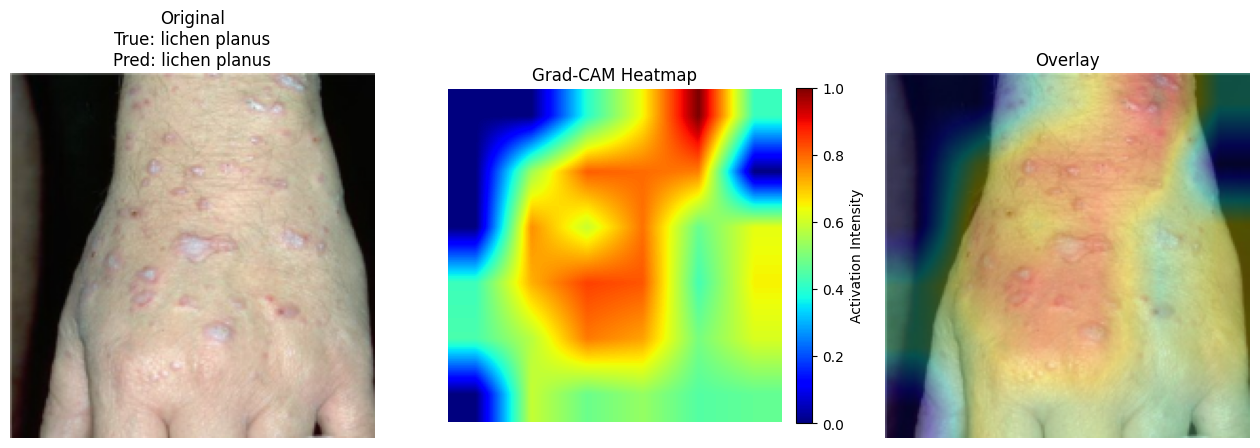}
        \caption{Lichen Planus.}
        \label{fig: Lichen Planus}
    \end{subfigure}
    \hfill
    % ----------- Third Image -----------
    \begin{subfigure}[b]{0.9\textwidth}
        \centering
        \includegraphics[width=\linewidth]{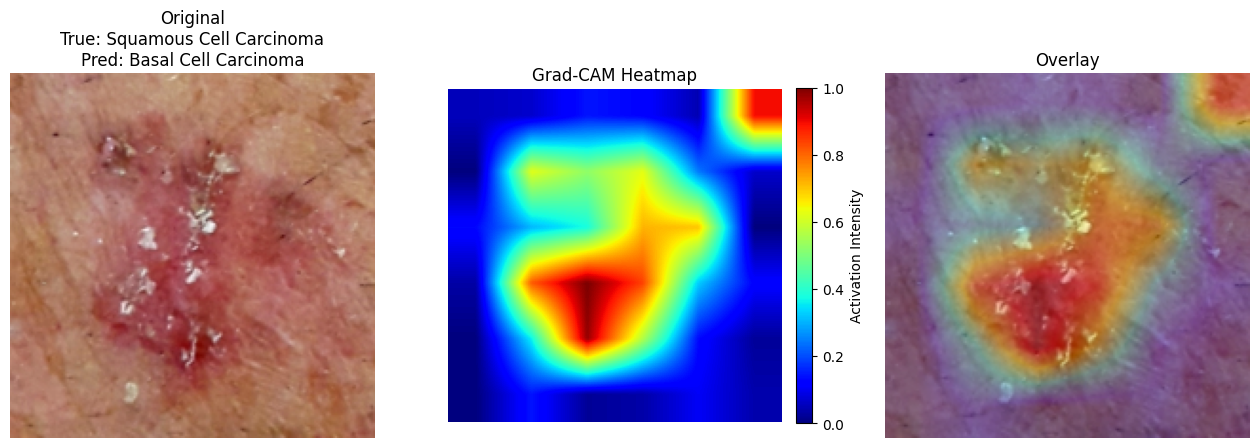}
        \caption{Squamous Cell Carcinoma.}
        \label{fig: Squamous Cell Carcinoma}
    \end{subfigure}
    
    \caption{Grad-CAM visualizations for representative cases. Each row shows the 
    original image, the Grad-CAM heatmap, and the overlay in a single composite figure. 
    Warmer regions (red/yellow) indicate higher contribution to the prediction, while 
    cooler regions (blue) indicate lower contribution.}
    \label{fig: gradcam_examples}
\end{figure}

Figure~\ref{fig: gradcam_examples} illustrates several representative Grad-CAM outputs. Each example consists of the original input image, the Grad-CAM heatmap, and the overlay of the heatmap on the input. Warmer regions (red/yellow) indicate areas of high importance, whereas cooler regions (blue) correspond to areas that contributed little to the prediction.

In correctly classified cases (e.g., Hand-Foot-and-Mouth Disease and Lichen Planus), the highlighted regions align well with lesion-affected skin patches, confirming that the model attends to clinically relevant features. 

In the misclassified case (Figure \ref{fig: Squamous Cell Carcinoma}), SCC was predicted as BCC. The Grad-CAM heatmap indicates that the model did focus on the lesion region. However, the prediction was still incorrect, likely due to the high degree of visual similarity between these conditions. Both classes share overlapping morphological features, such as irregular red patches and scaly textures \cite{calder2024bcc_vs_scc}, which can confound the model. This suggests that misclassification is not always a result of ignoring relevant features but rather of insufficient discriminative power between visually similar categories.

\subsection{Misclassification Analysis}

To gain deeper insight into model limitations and understand why the model is failing in certain cases, we examined errors from three complementary views: class-wise confusion matrix, LIME, and heatmaps from Grad-CAM. Together, these analyses reveal two dominant patterns:

\begin{enumerate}
    \item Fine-grained visual similarity: The most frequent confusions involved BCC, SCC, and AK. The best performing Swin Transformer misclassified 35 samples of BCC as AK and 29 as SCC (Figure \ref{fig:conf_b}). Confusion matrices showed reciprocal mislabeling among these classes; Grad-CAM indicated that the models often focused on the correct lesion region, yet class evidence was insufficiently distinctive. This suggests limited separability at the image level—clinically plausible given overlapping erythema, keratotic crusts, and irregular pigmentation in early presentations. It is demonstrated in Figure \ref{fig: BCC_SCC}.

\begin{figure}[htbp]
    \centering
    \includegraphics[width=\linewidth]{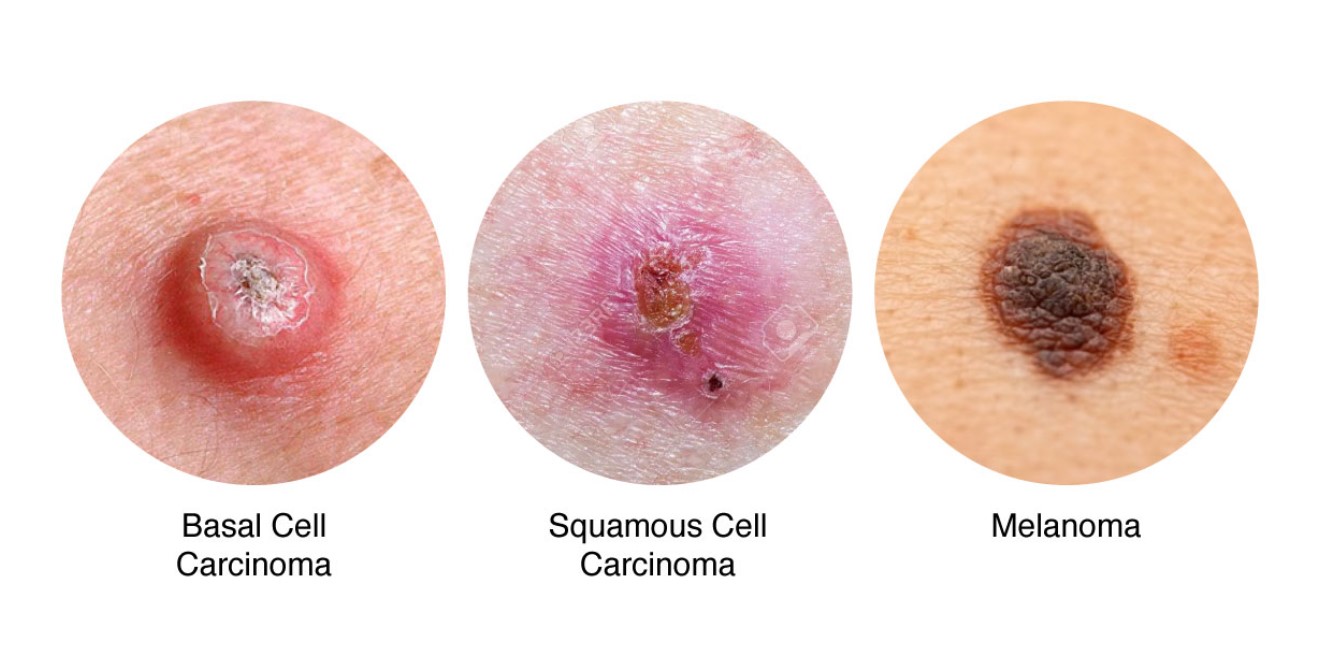}
    \caption{Visual similarity between BCC and SCC. Adapted from ~\cite{visage2025skincancer}.}
    \label{fig: BCC_SCC}
\end{figure}

    \item Acquisition-driven errors. Several false negatives (e.g., arsenic → normal) were linked to suboptimal capture: off-center framing, camera flash, motion blur, or low contrast (Figure \ref{fig:lime_misclassified}). LIME maps emphasized non-lesional skin folds or background textures, while Grad-CAM produced diffuse attention, indicating a weak signal. These cases highlight the need for simple capture protocols and an image-quality/ROI gate before classification.
    
\end{enumerate}

\begin{figure}[htbp]
    \centering
    % ----------- First Image -----------
    \begin{subfigure}[b]{0.95\textwidth}
        \centering
        \includegraphics[width=\linewidth]{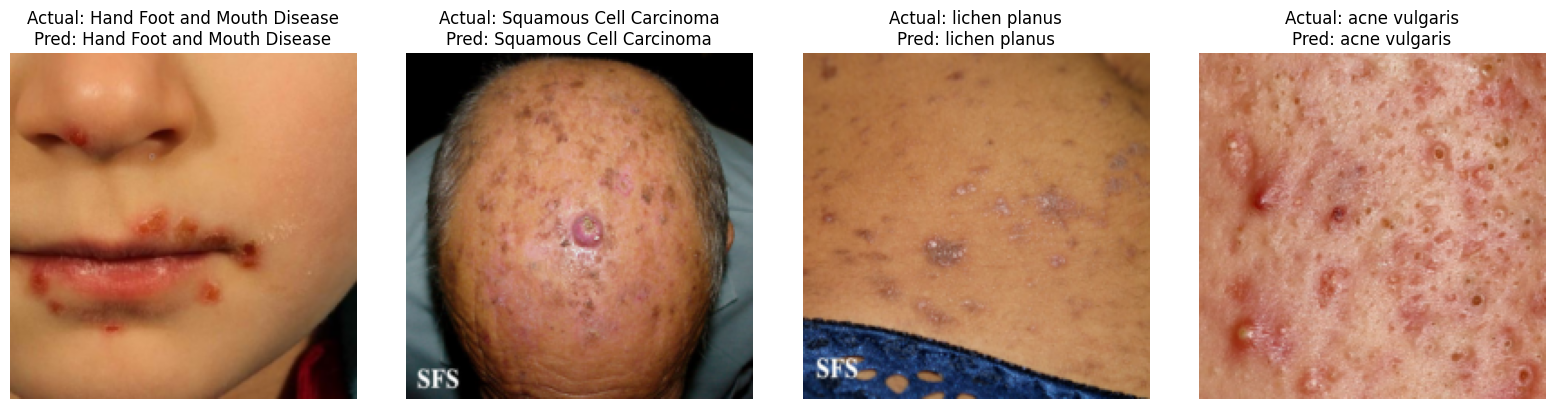}
        \caption{Correctly classified samples.}
        \label{fig: correct_01}
    \end{subfigure}
    \hfill
    % ----------- Second Image -----------
    \begin{subfigure}[b]{0.95\textwidth}
        \centering
        \includegraphics[width=\linewidth]{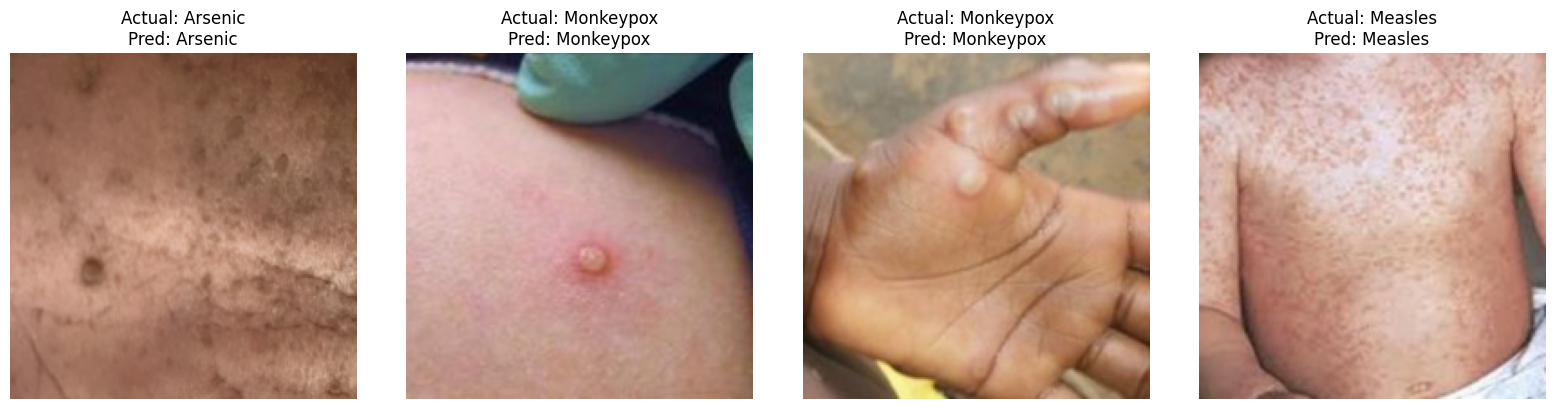}
        \caption{Correctly classified samples.}
        \label{fig: correct_02}
    \end{subfigure}
    \hfill
    % ----------- Third Image -----------
    \begin{subfigure}[b]{0.95\textwidth}
        \centering
        \includegraphics[width=\linewidth]{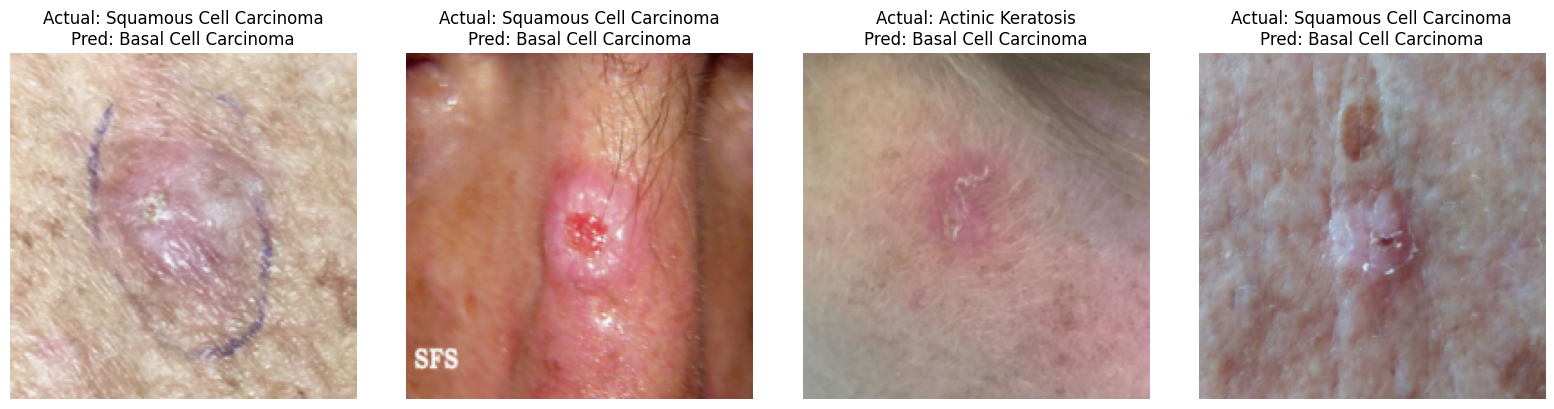}
        \caption{Misclassified samples.}
        \label{fig: incorrect_01}
    \end{subfigure}
    
    \caption{Model Predictions.}
    \label{fig: prediction_examples}
\end{figure}

To contextualize these findings, representative correctly classified and misclassified examples are shown in Figure~\ref{fig: prediction_examples}. As illustrated in Figures~\ref{fig: correct_01} and~\ref{fig: correct_02}, the correctly classified cases display lesions that are clearly visible with distinct morphological features, enabling the model to make reliable predictions. In contrast, the misclassified cases (Figure \ref{fig: incorrect_01}) often involve lesions that exhibit a high degree of visual similarity to other conditions, making differentiation challenging even when the model’s attention is focused on the relevant region.

To summarize, (i) Many errors are not due to “looking in the wrong place,” but to insufficient discriminative features between clinically similar classes; (ii) several are data-quality issues that can be mitigated by standardized capture and automated ROI detection/segmentation. Low-quality images can always appear in real-world settings. In such cases, looking at LIME- or Grad-CAM–generated heatmaps can provide valuable diagnostic support by confirming whether the model is still attending to lesion-specific regions or instead focusing on irrelevant artifacts. This additional layer of interpretability can help clinicians and health workers assess the reliability of model predictions and ultimately make more informed decisions in rural or resource-limited environments.

\subsection{Arsenic Case Analysis}

\begin{figure}[htbp]
    \centering
    % First misclassification
    \begin{subfigure}[b]{0.45\textwidth}
        \centering
        \includegraphics[width=\linewidth]{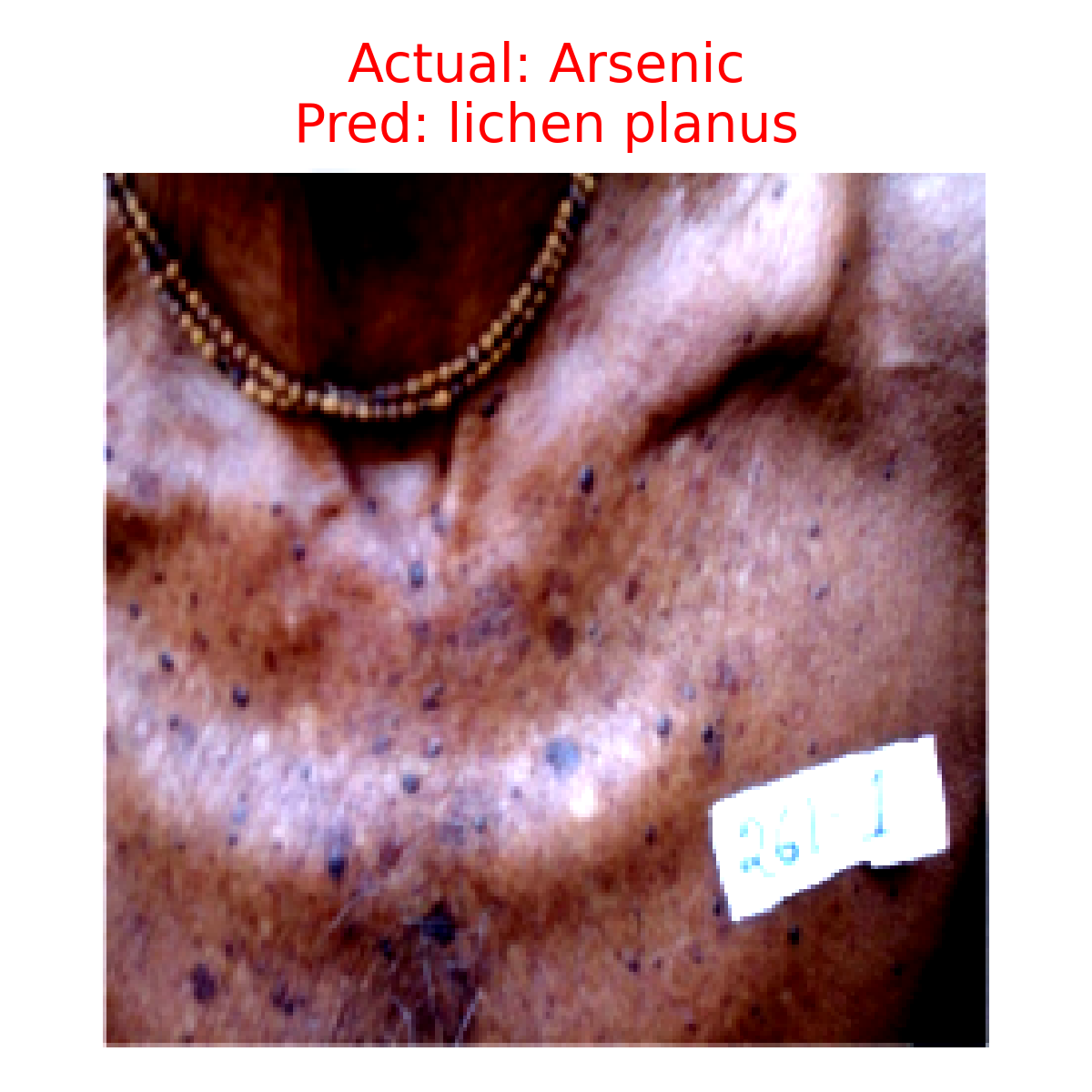}
        \caption{Actual: Arsenic, Pred: Lichen planus}
        \label{Fig: arsenic-> lichen}
    \end{subfigure}
    \hfill
    % Second misclassification
    \begin{subfigure}[b]{0.45\textwidth}
        \centering
        \includegraphics[width=\linewidth]{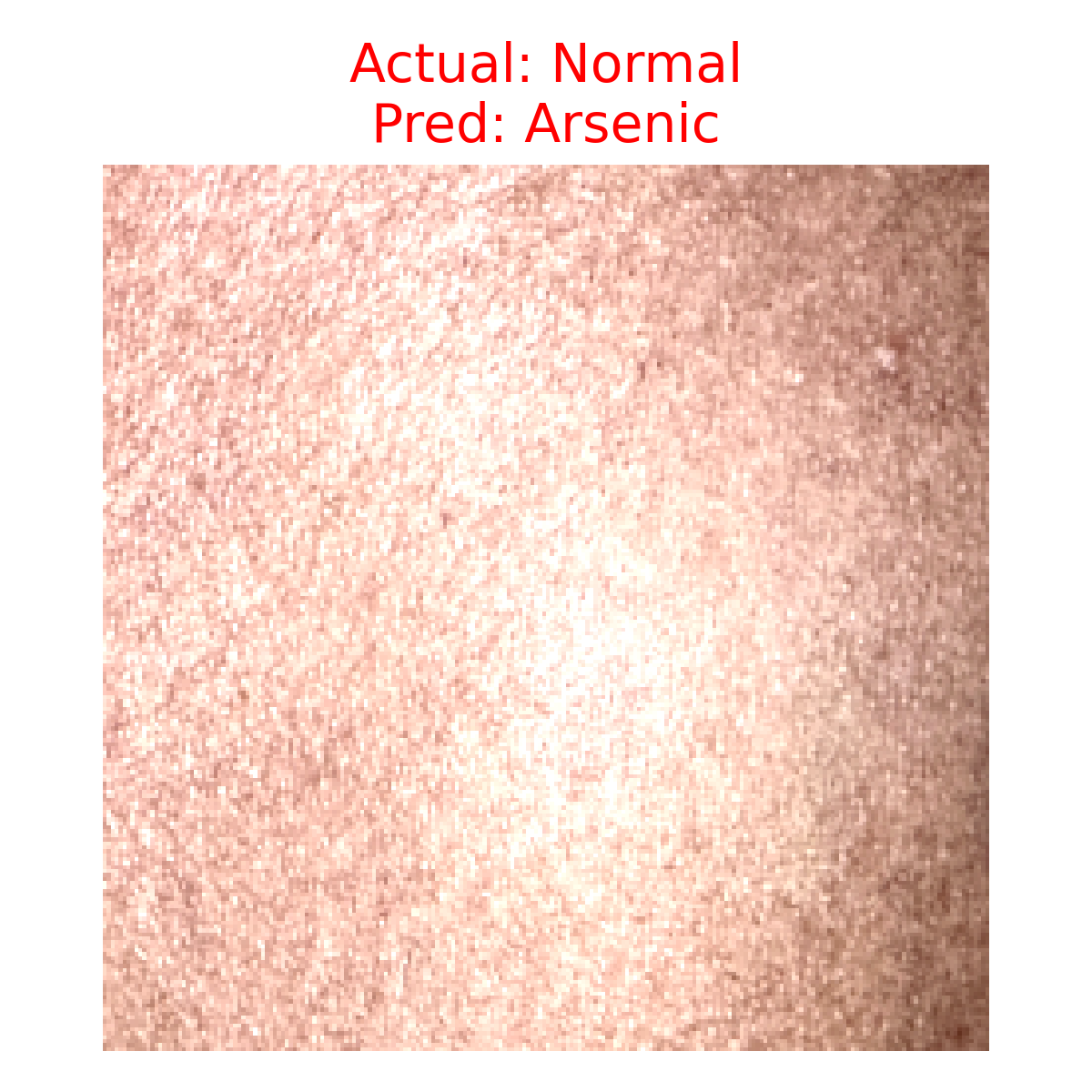}
        \caption{Actual: Normal, Pred: Arsenic}
        \label{Fig: normal-> arsenic}
    \end{subfigure}
    
    \caption{Representative misclassified examples. (a) An arsenic case misclassified as lichen planus. (b) A normal case misclassified as arsenic.}
    \label{fig:arsenic_misclassified}
\end{figure}

In this section, we look particularly into Arsenic-affected skin lesion detection. The confusion matrices reveal important trends regarding arsenic classification. While the Swin transformer almost perfectly identified all arsenic cases, other DL models occasionally 
made mistakes. Studying these errors can help us identify possible limitations of our diagnostic framework. Two representative misclassified cases are presented in Figure \ref{fig:arsenic_misclassified}.

In general, Arsenic lesions were most often misclassified as normal skin, hand-foot-and-mouth disease, monkeypox, and lichen planus (Figure \ref{Fig: arsenic-> lichen}). Many of these errors can be attributed to either the subtle presentation of early-stage arsenicosis (leading to confusion with normal) or strong visual overlap with other dermatoses, such as keratotic lesions (lichen planus), or poor image quality. The occasional misclassification as hand-foot-and-mouth disease or monkeypox likely arises from shared reddish or vesicular patterns, which the models found difficult to disentangle in low-quality or poorly captured images.

Conversely, several non-arsenic conditions were incorrectly predicted as arsenic. The most common among these were lichen planus, monkeypox, and normal cases. Lichen planus shares scaly plaques and pigmentation changes that can mimic arsenic keratosis, while monkeypox lesions can sometimes resemble hyperkeratotic papules. 

As for normal skin images, manual inspection of the misclassified samples revealed that in some cases, rough or uneven skin texture in otherwise normal images was misdiagnosed as arsenic (Figure \ref{Fig: normal-> arsenic}). These errors suggest that the model may be overly sensitive to coarse surface patterns and pigmentation irregularities, mistaking benign variations in skin texture for pathological keratosis. This is likely due to the presence of rough-textured or uneven normal skin images in the training data, which may have biased the model toward interpreting coarse surface patterns as arsenic-related keratosis. Careful curation of the normal class and inclusion of a wider variety of benign skin textures could help mitigate this issue.

These bidirectional confusions highlight two challenges: (i) the high visual similarity between arsenic-induced lesions and certain inflammatory or keratotic disorders, and (ii) the sensitivity of the models to image quality and capture conditions. While the overall arsenic classification accuracy was strong, these patterns underscore the need for better image quality and potentially the integration of clinical metadata to reduce false positives and false negatives.

\subsection{External Validation}

To further examine the robustness of the proposed framework, we performed a small-scale external validation using images outside the training and test distributions. These included samples from different classes. The model demonstrated strong generalization capability, correctly classifying the majority of cases (over 90\%).

\begin{figure}[htbp]
    \centering
    % Row 1
    \begin{subfigure}[b]{0.45\textwidth}
        \centering
        \includegraphics[width=\linewidth]{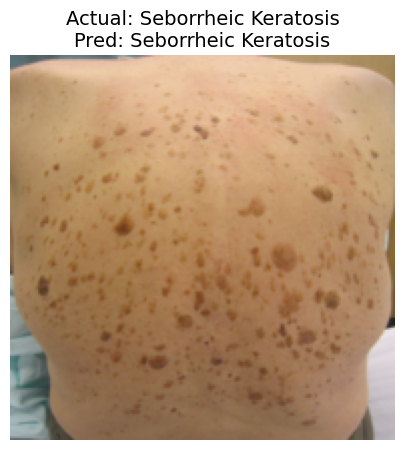}
        \caption{Actual: Seborrheic Keratosis \\ Pred: Seborrheic Keratosis}
    \end{subfigure}
    \hfill
    \begin{subfigure}[b]{0.45\textwidth}
        \centering
        \includegraphics[width=\linewidth]{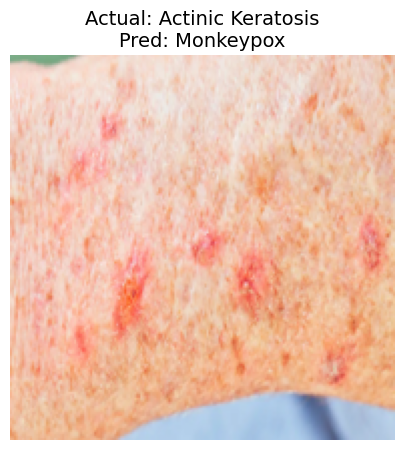}
        \caption{Actual: Actinic Keratosis \\ Pred: Monkeypox}
    \end{subfigure}
    
    \vspace{0.4cm} % space between rows
    
    % Row 2
    \begin{subfigure}[b]{0.45\textwidth}
        \centering
        \includegraphics[width=\linewidth]{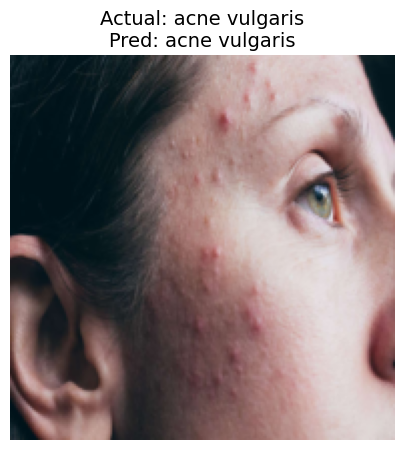}
        \caption{Actual: Acne Vulgaris \\ Pred: Acne Vulgaris}
    \end{subfigure}
    \hfill
    \begin{subfigure}[b]{0.45\textwidth}
        \centering
        \includegraphics[width=\linewidth]{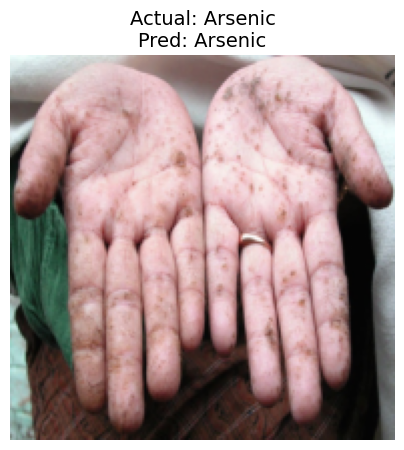}
        \caption{Actual: Arsenic \\ Pred: Arsenic}
    \end{subfigure}
    
    \caption{Representative external validation examples across multiple conditions.}
    \label{fig:external_validation}
\end{figure}

Figure~\ref{fig:external_validation} presents four representative examples (other examples are provided in the GitHub repository). The seborrheic keratosis and acne vulgaris cases were classified correctly with high confidence, indicating that the model generalizes well to common dermatoses. The arsenic case was also predicted correctly, showing its ability to capture arsenicosis-specific patterns even on unseen data. One limitation observed was the misclassification of an AK sample as monkeypox, which may be due to overlapping erythematous and scaly features. Overall, these results demonstrate the potential of the model to extend beyond curated datasets and perform effectively in real-world clinical scenarios.

\subsection{Web-Based Deployment Framework}

A web-based application was developed using a server-side rendered (SSR) Flask framework. The trained model was first exported into a deployable format (e.g., a .keras file for TensorFlow/Keras models or a .pth file for PyTorch models). This serialized model was then loaded on the server to handle real-time inference requests. Both prediction computation and LIME-based interpretability analysis are performed on the server side, and the processed outputs are rendered dynamically to the user interface. Within this framework, users can upload images, which are preprocessed on the server, passed through the model, and returned with both prediction scores and explanation overlays. This lightweight proof-of-concept application demonstrates not only the effectiveness of the proposed model in classifying arsenic-induced and other skin lesions but also its capability to provide interpretable visual explanations for each prediction.

%We adopted a server-side deployment strategy because it enables centralized model updates, reduces the computational burden on end-user devices, and ensures consistent performance across a wide range of hardware. This approach is particularly beneficial in resource-limited rural settings, where smartphones or computers may not have sufficient power to run heavy DL models locally.

As part of our future work, we plan to extend this framework by developing a lightweight mobile application that can perform on-device inference. This will enable fully offline operation, making the system usable in rural arsenic-affected communities where internet connectivity is unreliable or unavailable.

A short video demonstration of the web application has also been provided in the accompanying GitHub repository, allowing readers to visualize the workflow and interaction process.

\subsection{Data Augmentation}

Our curated dataset contained a limited number of samples for different categories. Even for the arsenic case, fewer than 1000 samples were collected. To increase the dataset size, we experimented with both conventional data augmentation techniques and Generative Adversarial Network (GAN)-based synthetic image generation. Standard augmentation techniques (e.g., rotations, flips, color adjustments) provided little measurable benefit. 

\begin{figure}[htbp]
    \centering
    % ---------- First image ----------
    \begin{subfigure}[b]{0.45\textwidth}
        \centering
        \includegraphics[width=\linewidth]{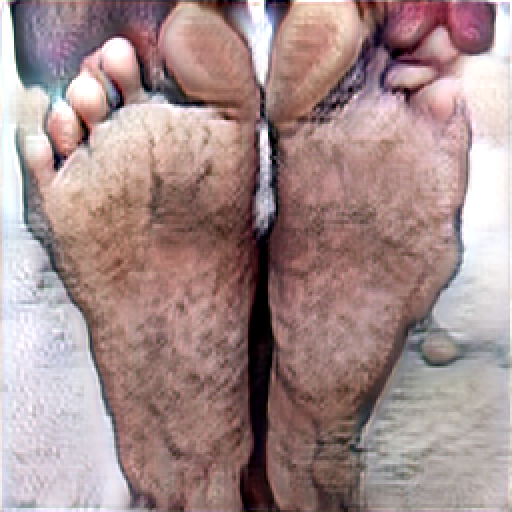} % replace with your first image
        \caption{GAN-generated arsenic image 1}
        \label{fig:gan_a}
    \end{subfigure}
    \hfill
    % ---------- Second image ----------
    \begin{subfigure}[b]{0.45\textwidth}
        \centering
        \includegraphics[width=\linewidth]{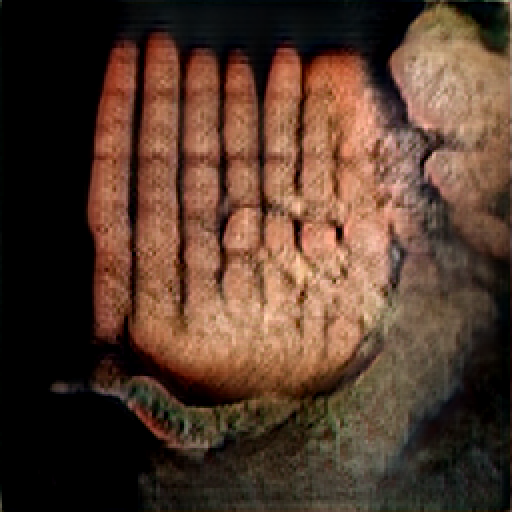} % replace with your second image
        \caption{GAN-generated arsenic image 2}
        \label{fig:gan_b}
    \end{subfigure}
    
    \caption{Examples of GAN-generated arsenic-affected skin images. 
    Both samples display unrealistic textures and distorted anatomical 
    structures, illustrating the limitations of GAN-based augmentation 
    for this dataset.}
    \label{fig:gan_examples}
\end{figure}

We also tried to apply GANs to synthesize additional training images for underrepresented classes, especially Arsenic. However, the quality of the generated samples was poor, as shown in Figure~\ref{fig:gan_examples}. The synthetic images frequently displayed unrealistic textures, color artifacts, and distorted anatomical structures, limiting their clinical usability.

This outcome is likely due to the relatively small number of training samples available. GANs are known to require large datasets to effectively capture complex visual patterns and generate realistic images. When trained on limited medical data, they often produce unstable or low-quality outputs. Similar challenges have been reported in the literature, where GANs trained on small-scale medical datasets struggled to generalize and produced visually implausible results~\cite{frid2018gan}. These low-quality samples failed to provide meaningful variability and risked introducing noise into the training process. Therefore, GAN-generated images were not included in the training process.

\subsection{Fine-Tuning}

In order to assess whether deeper adaptation of pretrained models could yield performance gains, we explored progressive fine-tuning strategies. During the initial training phase (first 10–15 epochs), the backbone weights were frozen and only the custom classification layers were optimized. This resulted in steady improvements in both training and validation performance for most architectures. However, once the final backbone layers were unfrozen and training resumed, certain models, such as InceptionV3 or VGG16, exhibited signs of overfitting: training accuracy continued to rise, while validation accuracy plateaued and validation loss diverged. For other models, including EfficientNetB0, the performance differences after fine-tuning were minimal and largely negligible.

These findings suggest that, under the constraints of a relatively small dataset, unfreezing large portions of the backbone may destabilize pretrained feature representations rather than improve them. The limited number of samples per category—some with fewer than a hundred images—makes it difficult to effectively update the millions of parameters in these networks. Similar observations have been reported in medical imaging studies, where feature extraction with frozen backbones and lightweight classifier heads has often proven more effective than extensive fine-tuning~\cite{kim2022transfer}.

\subsection{Limitations}

Despite the encouraging results, this study has several limitations that should be acknowledged. First, the dataset size remains relatively limited. The number of samples available for certain categories was quite limited. Even for the Arsenic cases, the number of samples that we managed to collect was less than 1000. Although transfer learning mitigated some of these challenges, the restricted number of samples per class limits the model’s ability to fully generalize to unseen cases. Moreover, class imbalance persists, with certain conditions being more represented than others. This imbalance may bias the model toward more frequent categories.

Second, the dataset was curated from mobile phone–captured images collected under real-world conditions. While this strengthens ecological validity, it also introduces variability in image quality, lighting, focus, and lesion framing, which in turn affects classification performance. As shown earlier, some misclassifications were directly attributable to poor image acquisition, underscoring the need for standardized capture protocols.

Third, all models relied on transfer learning from ImageNet-pretrained weights. These weights were derived from natural images rather than medical imagery, which may not optimally represent dermatological or arsenic-related features. A more tailored pretraining strategy (e.g., using large-scale dermatology datasets) could further improve performance.

Finally, this study is limited to image-based diagnosis. No clinical metadata (e.g., duration of exposure, comorbidities, or demographic information) was incorporated, although such information could substantially enhance diagnostic accuracy. In addition, the majority of arsenic-affected skin images were collected from Bangladesh, with no data from geographically diverse cohorts. This lack of population diversity restricts the generalizability of the findings to other regions and ethnic groups.

\section{Conclusion}

In this study, we developed and evaluated a DL–based framework for the automated classification of arsenic-induced skin lesions alongside a wide range of similar and common dermatological conditions. Leveraging a newly curated dataset of 20 classes of mobile phone–captured skin images, we benchmarked multiple state-of-the-art architectures, including both CNNs and Transformer-based models. The results demonstrate that while some CNNs achieved stable convergence and competitive accuracy, Transformer-based models, particularly the Swin Transformer, consistently outperformed all CNN variants, achieving the highest overall accuracy and MCC.

To improve model interpretability, we incorporated  XAI techniques (LIME and Grad-CAM), which provided insight into the regions most influential in the decision-making process. These visual explanations not only enhance clinical trust and transparency but also help identify misclassifications. Importantly, such interpretable outputs can support early diagnosis by providing healthcare workers with an additional layer of decision support.

External validation, performed on images outside the curated dataset, showed promising results, indicating strong potential for generalization beyond the curated dataset. Furthermore, the deployment of a lightweight web-based application demonstrates the feasibility of translating the proposed framework into a practical screening tool, especially in arsenic-affected rural communities with limited access to specialized healthcare.

Despite these promising outcomes, the study has several limitations, including the modest dataset size, class imbalance, variability in image quality, and lack of geographically diverse data. These limitations highlight the need for larger, more diverse, and clinically annotated datasets, as well as integration with clinical metadata to improve diagnostic robustness.

In conclusion, this work represents an important step toward AI-assisted, non-invasive, and accessible diagnosis of arsenicosis using mobile-acquired images. By enabling reliable image-based screening, it can serve as a practical diagnostic aid in rural and resource-limited communities, where access to dermatologists is scarce, thereby supporting early detection and timely intervention. 

Future research will focus on expanding the dataset, incorporating multimodal features (e.g., clinical history), and validating the models across diverse populations to ensure generalizability. In addition, we propose the design of a lightweight mobile application for on-device inference, which would be particularly valuable in rural regions with limited internet connectivity and healthcare access. Such efforts will be crucial for developing reliable, interpretable, and deployable diagnostic tools that can support early detection and health interventions in arsenic-affected regions around the world.

\vspace{1cm}

\noindent\textbf{Acknowledgments}

All code and experimental results associated with this study are openly available at: 
\url{https://github.com/newaz-aa/DL-Arsenicosis-Diagnosis}.

\bibliographystyle{elsarticle-num} 
\bibliography{arsenic_reference}

\end{document}